\documentclass[11pt]{article}

\usepackage[table,xcdraw]{xcolor}
\definecolor{NLcol}{HTML}{C04F15}
\definecolor{FOLcol}{HTML}{3B7D23}
\definecolor{MIXcol}{HTML}{1E77B4}
\definecolor{Finalcol}{HTML}{FF800E}
\definecolor{CMTgreen}{HTML}{228B22}
\definecolor{logiccol}{HTML}{FDE6D8}
\definecolor{affectcol}{HTML}{DDF2D8}
\definecolor{rhetoriccol}{HTML}{DCEBFA}
\usepackage{xcolor}
\newcommand{\downimp}[1]{\textcolor{green!45!black}{\scriptsize{(#1\%$\downarrow$)}}}
\newcommand{\upimp}[1]{\textcolor{green!45!black}{\scriptsize{(#1\%$\uparrow$)}}}
\newcommand{\gain}[1]{\textcolor{green!45!black}{\scriptsize{(+#1)}}}
\usepackage[most]{tcolorbox}

\newtcolorbox{findingbox}[1]{
  enhanced,
  breakable,
  colback=gray!4,
  colframe=gray!45!black,
  boxrule=0.4pt,
  arc=1.5pt,
  left=7pt,
  right=7pt,
  top=5pt,
  bottom=5pt,
  before skip=8pt,
  after skip=8pt,
  title=#1,
  coltitle=gray!35!black,
  fonttitle=\bfseries,
  attach title to upper,
  after title={\par\vspace{2pt}},
}
\usepackage[preprint]{acl}

\hypersetup{
    colorlinks=true,
    urlcolor=magenta,
    linkcolor=red
}

\usepackage{siunitx}
\usepackage{url}
\usepackage{amsthm}
\usepackage{booktabs}
\usepackage{multirow}
\usepackage{pifont}
\usepackage{listings}
\usepackage{xcolor}
\definecolor{codepurple}{rgb}{0.58,0,0.82}
\definecolor{codegray}{rgb}{0.5,0.5,0.5}
\definecolor{codeblue}{rgb}{0.0,0.2,0.6}
\definecolor{codered}{rgb}{0.8,0.0,0.0}
\definecolor{backcolour}{rgb}{0.98,0.98,0.98}
\usepackage[table]{xcolor}
\usepackage{multirow}
\usepackage{booktabs}
\usepackage{algorithm}
\usepackage{algorithmicx}
\usepackage{algpseudocode}
\usepackage{makecell}
\usepackage[most]{tcolorbox}
\usepackage{todonotes}

\usepackage{enumitem}
\usepackage{amssymb} 

\usepackage{algpseudocode}
\usepackage{subcaption}
\usepackage{colortbl}

\usepackage{amsthm}

\theoremstyle{definition}

\theoremstyle{plain}

\usepackage{times}
\usepackage{latexsym}

\usepackage[T1]{fontenc}

\usepackage[utf8]{inputenc}

\usepackage{microtype}

\usepackage{inconsolata}

\usepackage{graphicx}
\usetikzlibrary{positioning,arrows.meta,backgrounds,fit}
\usepackage{tikz}
\usetikzlibrary{shapes.geometric, arrows.meta, positioning, calc, chains, backgrounds, fit}
\usepackage{float} 

\definecolor{logiccol}{RGB}{232, 240, 254}
\definecolor{affectcol}{RGB}{252, 232, 230}
\definecolor{rhetoriccol}{RGB}{230, 244, 234}

\usepackage{graphicx}

\usepackage{graphicx}

\newcommand{\deepseek}{%
  \begin{tabular}[c]{@{}l@{}}
    \includegraphics[height=4ex]{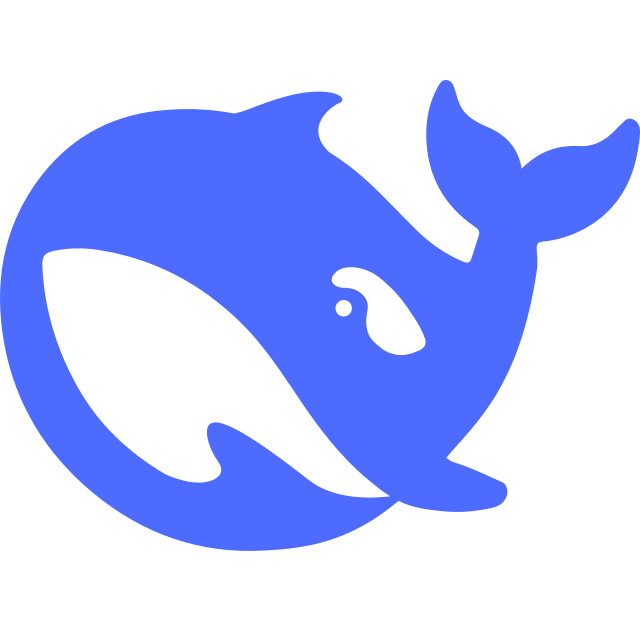} \\
    \scriptsize DeepSeek
  \end{tabular}%
}
\newcommand{\claude}{%
  \begin{tabular}[c]{@{}l@{}}
    \includegraphics[height=4ex]{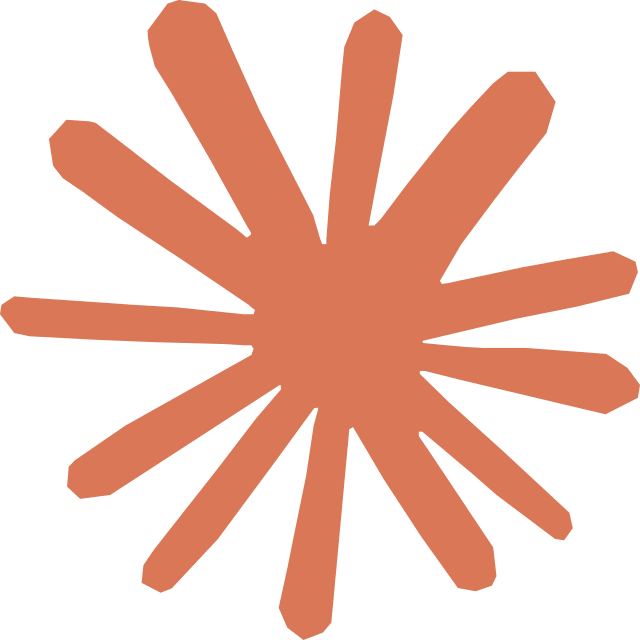} \\
    \scriptsize Claude
  \end{tabular}%
}
\newcommand{\gpt}{%
  \begin{tabular}[c]{@{}l@{}}
    \includegraphics[height=4ex]{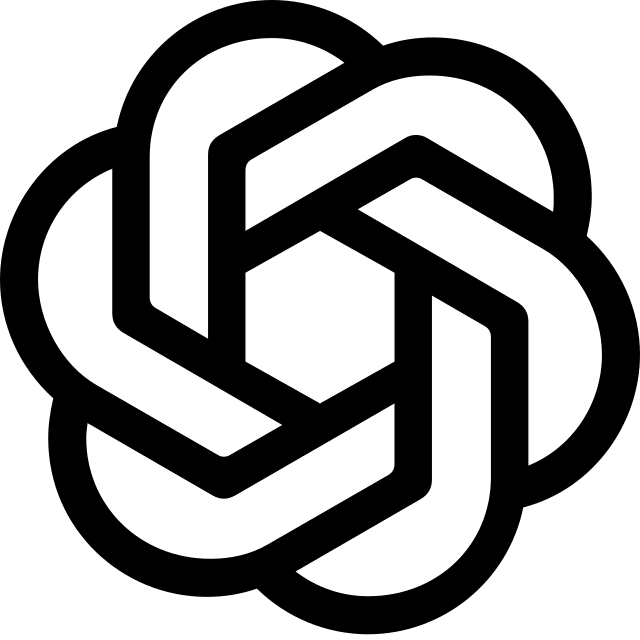} \\
    \scriptsize GPT
  \end{tabular}%
}
\newcommand{\llama}{%
  \begin{tabular}[c]{@{}l@{}}
    \includegraphics[height=4ex]{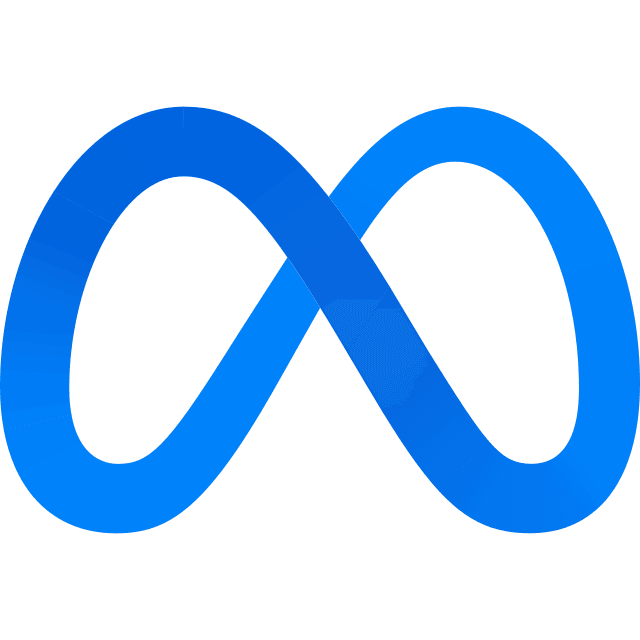} \\
    \scriptsize Llama
  \end{tabular}%
}
\usepackage{soul}
\sethlcolor{gray!20} 
\title{Too Good to Be Real? Diagnosing and Reducing the Gap Between AI Preference and Real User Engagement}

\author{
  \textbf{Xinglang Zhang},
  \textbf{Yuanmeng Xiang},
  \textbf{Yunyao Zhang},
  \textbf{Zeliang Chen},
  \textbf{Junqing Yu},
  \textbf{Zikai Song\thanks{Corresponding author.}}
  \\
  Huazhong University of Science and Technology
  \\
  \small{\{normanspark, skyesong\}@hust.edu.cn}
}

\begin{document}

\maketitle

\begin{abstract}
Large language models are increasingly used to generate and evaluate online content, yet it remains unclear whether the qualities they associate with higher engagement match what real users respond to.
We study this question using 1.17 million answers to 25,978 questions from Zhihu, Quora, and Reddit, comparing real platform answers and AI-generated answers across four within-question engagement levels.
We introduce \textbf{Ontological Preference Measurement}, which represents answers along three dimensions: logic, affect, and expression.
We find a systematic gap between \textbf{AI preference} and \textbf{real user engagement}: as target engagement increases, LLMs add more explicit logical structure, while real user engagement is more strongly associated with affective and expressive salience.
We call this tendency \textbf{logic overbinding}.
Based on this diagnosis, we propose \textbf{Ontology-Masked Reasoning Autoencoding (OMRA)}, a controlled intervention that masks and reconstructs over-explained spans while preserving stance, factual content, and coherence.
Across four LLM families, OMRA reduces the measured gap by an average of \textbf{54.4\%}.
In human evaluation, OMRA wins \textbf{67.4\%} of pairwise preference judgments against matched real platform answers, even though the real answers are more often judged to be human-written.
\end{abstract}


\section{Introduction}

\begin{figure}[!t]
  \centering
  \includegraphics[
    width=0.78\linewidth,
    keepaspectratio
  ]{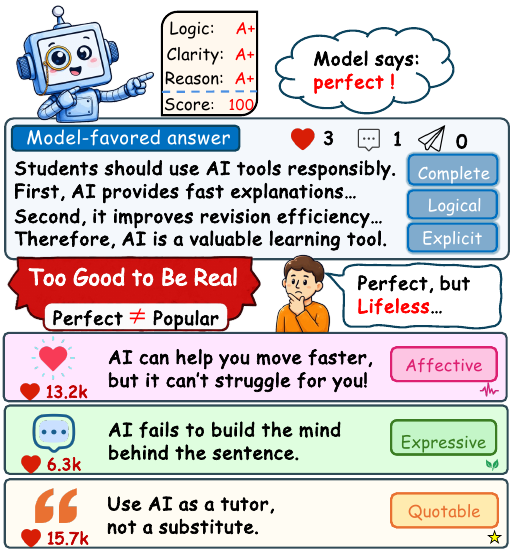}
  \caption{
\textbf{Illustrative example of the AI preference--user engagement gap.}
The AI-preferred answer is complete and explicitly reasoned but receives low engagement.
Answers with stronger affective or expressive salience receive higher engagement in this example.
}
  \label{fig:teaser}
\end{figure}

Large language models are increasingly used to generate, rank, and evaluate online content, and can be prompted to write answers targeting different levels of user engagement~\cite{ouyang2022training,lambert2025rewardbench}.
However, it remains unclear whether the changes they make reflect what real users actually respond to.
When targeting higher engagement, LLMs tend to make their answers more complete, more structured, and more explicitly reasoned~\cite{rodrigues2026linguistic}.
Real high-engagement answers, however, may follow different patterns.
This raises two questions:
\emph{What do LLMs change when asked to write an answer for higher user engagement?}
\emph{Do these changes match the patterns found in real high-engagement answers?}

To answer these questions, we compare real platform answers and AI-generated answers across the same four engagement levels.
We collect 1.17 million answers to 25,978 questions from Zhihu, Quora, and Reddit.
Because raw vote counts are affected by exposure, platform traffic, and time, we rank answers only against other answers to the same question and map them to four \textbf{within-question engagement levels}.
We then ask LLMs to generate an answer for each \textbf{target engagement level}.
This setup allows us to compare how real answers vary with \textbf{real user engagement} and how LLMs modify their answers when targeting the same levels.
We refer to the latter pattern as \textbf{AI preference}: the qualities that LLMs associate with higher engagement.

To compare these patterns, we introduce \textbf{Ontological Preference Measurement}, which represents each answer along three dimensions: \textbf{logic}, \textbf{affect}, and \textbf{expression}.
Logic captures explicit claims and support; affect captures emotional and value-oriented appeal; and expression captures rhetorical salience and memorability.
As target engagement increases, LLMs add more claims, evidence, causal explanations, and step-by-step justification, whereas real user engagement is more strongly associated with affective and expressive salience and only weakly negatively associated with explicit reasoning.
We call this tendency \textbf{logic overbinding}: AI preference places disproportionate weight on explicit logical presentation relative to real user engagement.
This pattern is consistent across transfer, marginal, and geometric comparisons.

Based on this diagnosis, we propose \textbf{Ontology-Masked Reasoning Autoencoding (OMRA)}, a controlled intervention that masks and reconstructs over-explained spans while preserving stance, factual content, and coherence.
Across four LLM families, OMRA reduces the measured AI preference--user engagement gap by an average of \textbf{54.4\%}, outperforming few-shot imitation, reasoning-augmented generation, and alternative masking strategies.
Additional controls show that length contributes to the gap but does not explain OMRA's improvement, while sentence-mood changes have only minor effects; preservation checks indicate that stance, factual consistency, relevance, and usefulness are largely retained.
In human evaluation, OMRA wins \textbf{67.4\%} of pairwise preference judgments against matched real platform answers, even though the real answers are more often judged to be human-written.

In summary, our contributions are:
\begin{itemize}[leftmargin=10pt, topsep=2pt, itemsep=0pt, label=\textbullet]
    \item We introduce \textbf{Ontological Preference Measurement}, a structured framework for comparing AI preference and real user engagement across logic, affect, and expression, and identify \textbf{logic overbinding} using 1.17 million answers from three platforms.
    
    \item We propose \textbf{Ontology-Masked Reasoning Autoencoding (OMRA)}, which reconstructs over-explained spans and reduces the measured gap by 54.4\% across four LLM families, with supporting evidence from control experiments and human evaluation.
\end{itemize}

\begin{figure*}[t]
\centering
\includegraphics[width=\linewidth, height=0.58\linewidth]{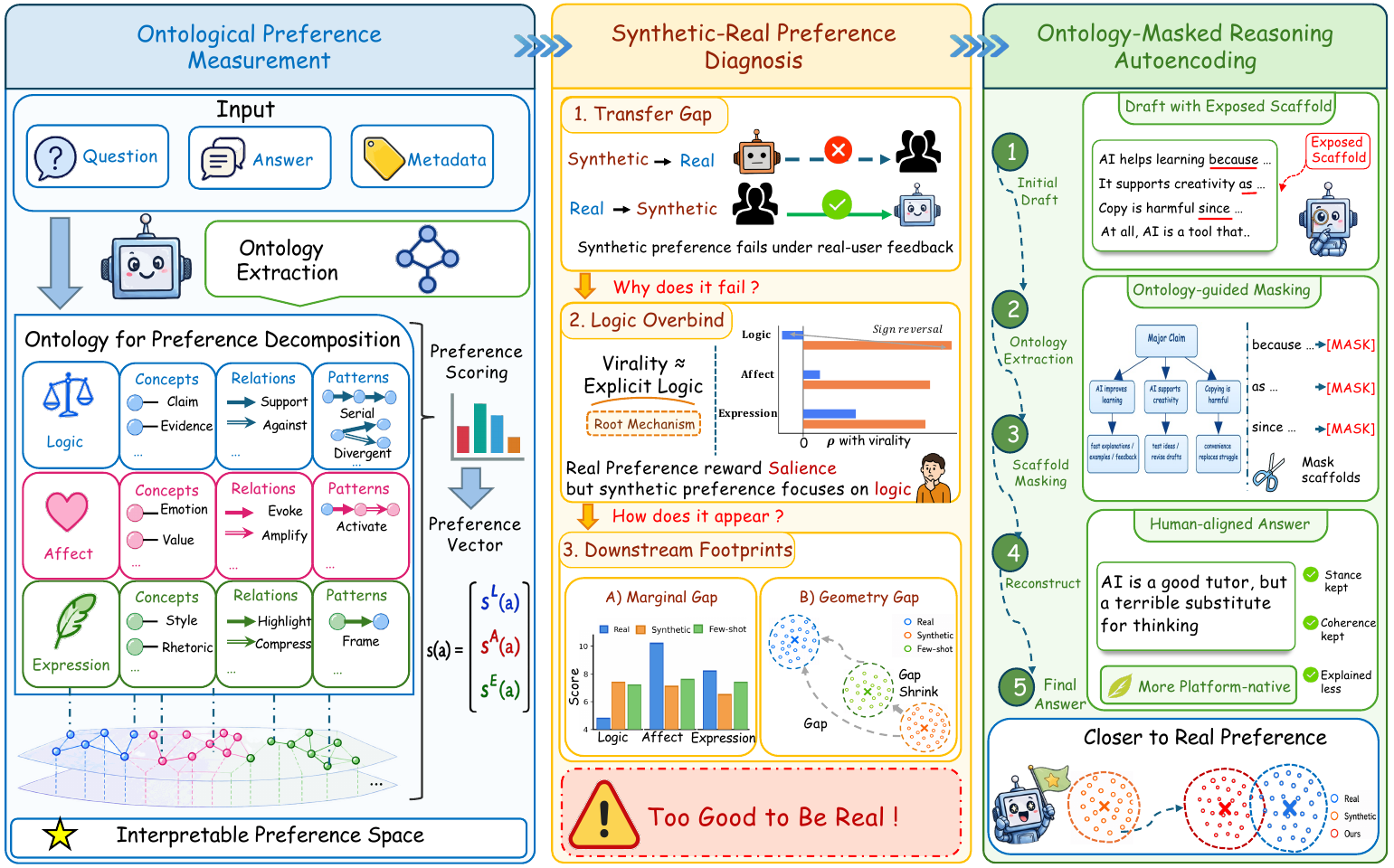}
\caption{
\textbf{Overview of our ontology-guided diagnosis and correction framework.}
\textbf{Left: Ontological Preference Measurement.} Answers are represented along logic, affect, and expression in a shared interpretable space.
\textbf{Middle: AI preference--user engagement diagnosis.} Transfer, marginal, and geometric comparisons reveal logic overbinding in AI preference.
\textbf{Right: Ontology-Masked Reasoning Autoencoding (OMRA).} OMRA identifies, masks, and reconstructs over-explained spans, moving AI-generated answers closer to the real-answer distribution.
}
\label{fig:pipeline}
\end{figure*}

\section{Related Work}
\label{sec:related_work}
A full discussion appears in Appendix~\ref{app:related_work} and we only highlight key directions here.

\paragraph{Ontology, discourse, and argument mining.}
Ontology-based modeling represents text through interpretable concepts and relations~\cite{gruber1993translation,noy2001ontology}, while argument mining identifies claims, evidence, and argumentative relations~\cite{lippi2016argumentation,lawrence2019argument,chakrabarty2019ampersand}. Related work also studies how discourse structure, affect, framing, and rhetoric shape persuasion and user response~\cite{feng2011classifying,tan2016winning,stab2017parsing}. We build on these directions by organizing online answers along logic, affect, and expression to compare AI preference with real user engagement in a shared interpretable space.

\paragraph{AI preference and LLM-as-a-judge.}
AI synthetic data and LLM-as-a-judge evaluation are widely used as scalable substitutes for human annotation in reward modeling and alignment~\cite{christiano2017deep,ouyang2022training,li2024crowdsourced,lambert2025rewardbench}. Prior work documents biases involving verbosity, length, position, and self-preference~\cite{hu2024explaining,ye2025justice,saito2023verbosity,wang2025eliminating}, but these biases are typically evaluated against benchmark labels or other models. We instead compare the qualities associated with AI preference against naturally occurring real user engagement.

\paragraph{Engagement modeling and preference-controlled generation.}
Prior work predicts online engagement using user, temporal, network, and content features~\cite{deng2020contributes,rameez2022viralbert,lei2025godbench,wang2026seeing}, while preference-aligned and controllable generation steer outputs toward desired rewards or attributes~\cite{ouyang2022training,rafailov2023direct,zhang2023survey,song2026socialintelligence,chen2026show}. These approaches do not directly diagnose which answer qualities AI preference associates with engagement or how those associations differ from real user engagement. OMRA addresses this gap by using the ontology to identify and reconstruct over-explained spans under stance-, factual-consistency-, and coherence-preserving constraints.

\section{Methodology}
\label{sec:methodology}

We organize our methodology into three stages illustrated in Figure~\ref{fig:pipeline}: ontological preference measurement (\S\ref{sec:measuring}), AI preference--user engagement gap diagnosis (\S\ref{sec:discovery}), and OMRA correction (\S\ref{sec:OMRA}).

\subsection{Ontological Preference Measurement}
\label{sec:measuring}

To study real user engagement beyond raw vote counts, we build an ontology-based measurement framework that represents each answer through three complementary dimensions---logic, affect, and expression. This framework connects observable platform engagement signals with interpretable textual properties rather than treating engagement as a black-box scalar.

\begin{figure*}[t]
    \centering
    \includegraphics[width=\textwidth,height=0.3\linewidth]{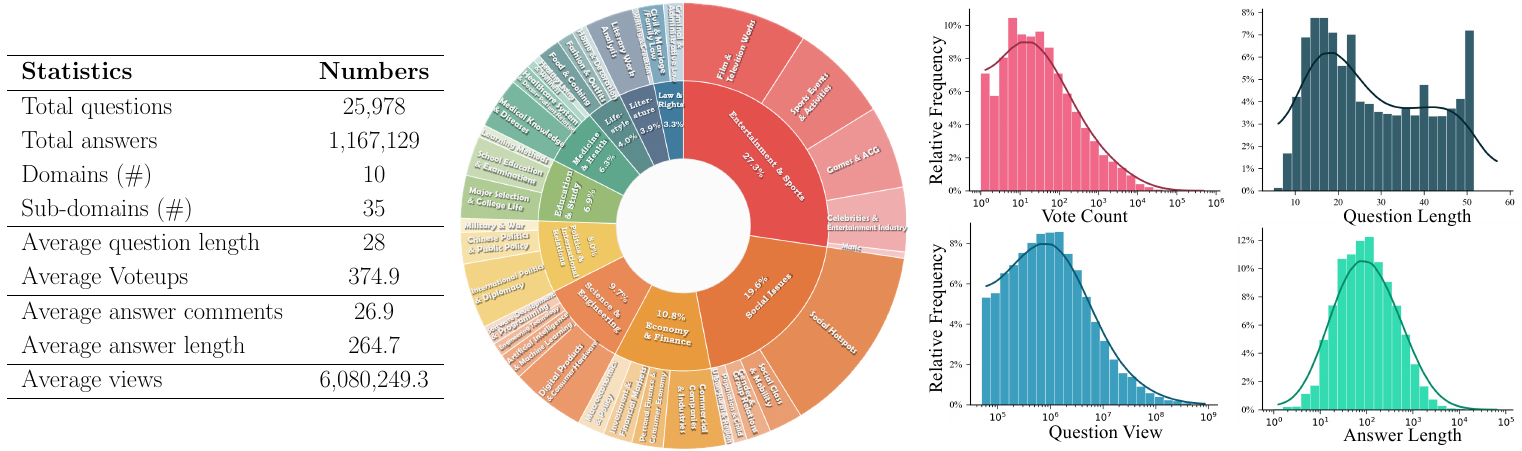}
    \caption{
    \textbf{Overview of the platform engagement dataset.}
    The left panel summarizes corpus-level statistics, including the number of questions, answers, domains, sub-domains, and engagement-related attributes.
    The middle sunburst chart shows the hierarchical topic distribution across primary domains and fine-grained sub-domains.
    The right panel visualizes the distributions of platform engagement signals and textual properties, including vote count, question title length, question view count, and answer length.
    }
    \label{fig:data_overview}
\end{figure*}

\subsubsection*{Real User Engagement Dataset}
\label{sec:dataset}

We collect a large-scale dataset of question--answer pairs with platform engagement signals from three open online platforms: \textbf{Zhihu}, \textbf{Quora}, and \textbf{Reddit}. The corpus spans 2020--2026 and covers 10 categories and 35 subcategories, with each answer associated with naturally occurring engagement signals such as votes (Figure~\ref{fig:data_overview}). Platform-specific filtering, deduplication, and engagement-quality criteria are detailed in Appendix~\ref{app:dataset}.

Since absolute vote counts are affected by question exposure, platform traffic, and temporal dynamics, we treat votes as \emph{platform engagement signals} rather than direct measurements of underlying human preference. We therefore operate on \emph{within-question engagement}. For each question $q$ with answer set $\mathcal{A}_q$, we assign each answer $a$ a \textbf{within-question engagement level}
\[
\ell(a)=\phi_q\!\left(\operatorname{rank}_q(v(a))\right)\in\{0,1,2,3\},
\]
where $\phi_q$ maps vote ranks within $q$ to four ordinal levels, isolating relative user engagement under a shared question context.

We compare four \emph{answer regimes}: \textbf{Real}, consisting of observed platform answers labeled by within-question engagement; \textbf{Synthetic}, consisting of AI-generated answers conditioned on target engagement levels; \textbf{Few-shot}, additionally conditioned on a real reference answer from the same level; and \textbf{OMRA-corrected}, revised by our ontology-guided masking and reconstruction method (\S\ref{sec:OMRA}).

\subsubsection*{Ontology for Preference Decomposition}
\label{sec:ontology}

We decompose each answer $a$ into three complementary preference layers,
\[
\mathcal{O}(a)=\bigl\{\mathcal{O}^{L}(a),\ \mathcal{O}^{A}(a),\ \mathcal{O}^{E}(a)\bigr\},
\]
where $\mathcal{O}^L$ (\textbf{logic}) captures how claims are argued and supported,
$\mathcal{O}^A$ (\textbf{affect}) captures emotional and value-oriented resonance,
and $\mathcal{O}^E$ (\textbf{expression}) captures rhetorical realization and salience.
This functional decomposition distinguishes complementary textual mechanisms
rather than collapsing them into a single quality score.
Each layer is organized around \emph{concepts}, \emph{relations}, and \emph{patterns},
forming a task-specific ontology of engagement-related textual cues rather than
an exhaustive account of engagement determinants.
Full ontology definitions, subdimensions, extraction templates, and examples
are provided in Appendix~\ref{app:ontology}.

\subsubsection*{Ontology Extraction and Preference Scoring}
\label{sec:scoring}

Given an answer, we use a fixed LLM-based extractor with a constrained ontology schema and output format to obtain $\mathcal{O}(a)$, and then convert it into a layer-wise preference vector
\[
\mathbf{s}(a)=\bigl[\mathbf{s}^{L}(a);\ \mathbf{s}^{A}(a);\ \mathbf{s}^{E}(a)\bigr],
\]
where each $\mathbf{s}^{\ast}(a)$ aggregates the scores of its subdimensions. The original answer text remains the primary evidence for scoring, while the extracted ontology provides structured cues. The resulting $\mathbf{s}(a)$ serves as an interpretable measurement of answer qualities rather than a direct engagement score, and grounds the diagnostics and corrections in the rest of the paper. Schemas and prompts are provided in Appendix~\ref{sec:full-prompting}.

\subsection{AI Preference--User Engagement Gap Diagnosis}
\label{sec:discovery}

Using the ontological preference vector $\mathbf{s}(a)$, we diagnose the
AI preference--user engagement gap.
We organize transfer, marginal, and geometric gaps as a diagnostic
sequence: from cross-regime transfer failure, to the association pattern
behind it, and finally to its distributional footprints.
Let $r \in \{R,S,F\}$ denote an answer regime
(Real, Synthetic, or Few-shot) and $\mathcal{D}_r$ its answer set.

\paragraph{Transfer Gap}
We first test whether an engagement rule learned in one regime
generalizes to another.
We train a predictor $f_r$ that maps
$\mathbf{s}(a)$ to within-question engagement level and evaluate it on
regime $r'$:
\begin{equation*}
T_{r \to r'} =
\mathcal{M}\!\left(f_r,\mathcal{D}_{r'}\right),
\qquad r,r' \in \{R,S,F\},
\end{equation*}
where $\mathcal{M}$ includes Spearman $\rho$ and Top-1 accuracy.
The asymmetry between $T_{R \to S}$ and $T_{S \to R}$ distinguishes
internal consistency from transfer: an AI preference rule may be
internally coherent yet fail on real platform answers.

\paragraph{Logic Overbinding}
We define \textbf{logic overbinding} as the tendency of AI preference to
place disproportionate weight on explicit logical presentation relative
to real user engagement.
For each $d \in \{L,A,E\}$, we compute the within-regime association
\begin{equation*}
\rho_d^{r} =
\operatorname{Spearman}\bigl(
\{s_d(a)\}_{a \in \mathcal{D}_r},
\{\ell(a)\}_{a \in \mathcal{D}_r}
\bigr),
\end{equation*}
where $L$, $A$, and $E$ denote logic, affect, and expression.
Comparing $\rho_d^{R}$ and $\rho_d^{S}$ tests whether AI preference and
real user engagement track the same answer qualities, or whether AI
preference more strongly favors explicit claim--evidence bridges, causal
justifications, and enumerated explanations.

\paragraph{Marginal Gap}
We compare marginal score distributions.
For $d \in \{L,A,E\}$, we compute 1-Wasserstein distance and
Jensen--Shannon divergence:
\begin{align*}
W_d^{r,r'} &=
W_1\bigl(\mathcal{P}_d^{r},\mathcal{P}_d^{r'}\bigr), \\
J_d^{r,r'} &=
\mathrm{JS}\bigl(\mathcal{P}_d^{r},\mathcal{P}_d^{r'}\bigr),
\end{align*}
where $\mathcal{P}_d^{r}$ is the empirical distribution of
$\{s_d(a)\}_{a \in \mathcal{D}_r}$.
While $\rho_d^{r}$ measures association with engagement,
$W_d^{r,r'}$ and $J_d^{r,r'}$ measure distributional differences across
answer regimes.

\paragraph{Geometric Gap}
Marginal similarity does not imply alignment in the joint preference
space: two regimes may overlap on individual dimensions while combining
them differently.
We therefore compare regimes in
$\mathbf{s}(a) \in \mathbb{R}^{|L|+|A|+|E|}$ using centroid distance and
maximum mean discrepancy:
\begin{alignat*}{2}
&C^{r,r'} &&=
\left\|
\mathbb{E}_{\mathcal{D}_r}[\mathbf{s}]
-
\mathbb{E}_{\mathcal{D}_{r'}}[\mathbf{s}]
\right\|_2, \\
&\operatorname{MMD}^{r,r'} &&=
\operatorname{MMD}\bigl(\mathcal{D}_r,\mathcal{D}_{r'}\bigr).
\end{alignat*}
Low-dimensional projections show how answer regimes occupy the shared
preference space.

\medskip
\noindent
Together, these probes provide complementary views of the same
AI preference--user engagement gap.
Section~\ref{sec:analysis} shows that their results consistently point
to logic overbinding and over-explained spans, motivating a targeted
intervention rather than separate optimization of each metric.

\subsection{Ontology-Masked Reasoning Autoencoding}
\label{sec:OMRA}

Based on this diagnosis, we propose
\textbf{Ontology-Masked Reasoning Autoencoding (OMRA)} as both a
controlled intervention and a method for reducing the measured
AI preference--user engagement gap.
OMRA targets over-explained spans that make reasoning unnecessarily
explicit, repetitive, or formulaic.

Inspired by masked autoencoding~\cite{he2022masked}, OMRA identifies and
masks ontology-derived over-explained spans, then reconstructs the answer
under stance-preserving, factual-consistency, coherence, and preference
constraints.
Given a Synthetic draft $a$ and its ontology representation
$\mathcal{O}(a)$ from \S\ref{sec:scoring}, OMRA produces a revised answer
$a'$ through the five stages in Algorithm~\ref{alg:OMRA}.

\begin{algorithm}[t]
\caption{\textbf{The OMRA Algorithm}}
\label{alg:OMRA}
\small
\begin{algorithmic}[1]
\Require Synthetic draft $a$; ontology extractor $\mathcal{E}$;
reconstructor $\mathcal{R}$; reviser $\mathcal{V}$;
preference constraints $c$; over-explanation type set
$\mathcal{T}_{\mathrm{over}}$
\vspace{4pt}

\State \textcolor{CMTgreen}{// Stage 1: Ontology Extraction}
\State $\mathcal{O}(a) \leftarrow \mathcal{E}(a)$, with
$\mathcal{O}(a)=
\{\mathcal{O}^L(a),\mathcal{O}^A(a),\mathcal{O}^E(a)\}$

\vspace{4pt}
\State \textcolor{CMTgreen}{// Stage 2: Over-Explained Span Identification}
\State $\mathcal{S}(a) \leftarrow
\bigl\{
\mathrm{span}(u)
\mid
u \in \mathcal{O}^L(a)\cup\mathcal{O}^E(a),
\mathrm{type}(u)\in\mathcal{T}_{\mathrm{over}}
\bigr\}$

\vspace{4pt}
\State \textcolor{CMTgreen}{// Stage 3: Ontology-Guided Masking}
\State $\tilde{a} \leftarrow \mathrm{Mask}(a,\mathcal{S}(a))$

\vspace{4pt}
\State \textcolor{CMTgreen}{// Stage 4: Reconstruction with Preference Constraints}
\State $\hat{a} \leftarrow
\mathcal{R}(\tilde{a},\mathcal{O}(a),c)$

\vspace{4pt}
\State \textcolor{CMTgreen}{// Stage 5: Preference-Aware Revision}
\State $a' \leftarrow
\mathcal{V}(\hat{a},\mathcal{O}(a),c)$

\vspace{4pt}
\State \textbf{Output:} revised answer $\textcolor{Finalcol}{a'}$
\end{algorithmic}
\end{algorithm}

\begin{table*}[t]
\centering
\scriptsize
\setlength{\tabcolsep}{4.2pt}
\renewcommand{\arraystretch}{1.08}
\begin{tabular}{llccccccc}
\toprule
\textbf{Base Model}
& \textbf{Method}
& \multicolumn{2}{c}{\textbf{Marginal Gap} $\downarrow$}
& \multicolumn{2}{c}{\textbf{Geometry Gap} $\downarrow$}
& \multicolumn{2}{c}{\textbf{Transfer to Real} $\uparrow$}
& \textbf{Avg. Shrink} $\uparrow$ \\
\cmidrule(lr){3-4}
\cmidrule(lr){5-6}
\cmidrule(lr){7-8}
&
& \textbf{Wass.}
& \textbf{JS}
& \textbf{MMD}
& \textbf{Cent.}
& \makecell{$\boldsymbol{\rho}$ \\ \scriptsize($\Delta$)}
& \textbf{Top-1}
& \\
\midrule
\multirow{5}{*}{\deepseek}
& Direct
& 1.479 & 0.154 & 0.240 & 1.969 & -0.051 & 0.261 & --- \\
& Few-shot
& 0.829 \downimp{43.9}
& 0.085 \downimp{44.8}
& 0.102 \downimp{57.5}
& 0.971 \downimp{50.7}
& 0.164 \gain{0.215}
& 0.316 \upimp{21.1}
& 49.2\% \\
& ToT
& 1.245 \downimp{15.9}
& 0.199 \textcolor{red}{($\uparrow$29.0\%)}
& 0.152 \downimp{36.8}
& 1.570 \downimp{20.3}
& 0.136 \gain{0.187}
& 0.313 \upimp{19.9}
& 11.0\% \\
& RoT
& 1.343 \downimp{9.2}
& 0.210 \textcolor{red}{($\uparrow$36.4\%)}
& 0.184 \downimp{23.1}
& 1.800 \downimp{8.6}
& -0.030 \gain{0.021}
& 0.253 \textcolor{red}{($\downarrow$3.1\%)}
& 1.1\% \\
& \textbf{Ours}
& \textbf{0.452} \downimp{69.4}
& \textbf{0.050} \downimp{67.5}
& \textbf{0.038} \downimp{84.2}
& \textbf{0.422} \downimp{78.6}
& \textbf{0.225} \gain{0.276}
& \textbf{0.358} \upimp{37.2}
& \textbf{74.9\%} \\
\midrule
\multirow{5}{*}{\claude}
& Direct
& 1.296 & 0.157 & 0.252 & 1.782 & -0.097 & 0.239 & --- \\
& Few-shot
& 0.987 \downimp{23.8}
& 0.135 \downimp{14.0}
& 0.138 \downimp{45.2}
& 1.197 \downimp{32.8}
& 0.074 \gain{0.171}
& 0.284 \upimp{18.8}
& 29.0\% \\
& ToT
& 1.427 \textcolor{red}{($\uparrow$10.1\%)}
& 0.227 \textcolor{red}{($\uparrow$44.5\%)}
& 0.176 \downimp{30.1}
& 1.670 \downimp{6.3}
& -0.047 \gain{0.050}
& 0.269 \upimp{12.6}
& \textcolor{red}{-4.6\%} \\
& RoT
& 1.571 \textcolor{red}{($\uparrow$21.2\%)}
& 0.237 \textcolor{red}{($\uparrow$50.8\%)}
& 0.212 \downimp{16.0}
& 1.854 \textcolor{red}{($\uparrow$4.0\%)}
& -0.023 \gain{0.074}
& 0.257 \upimp{7.5}
& \textcolor{red}{-15.0\%} \\
& \textbf{Ours}
& \textbf{0.807} \downimp{37.7}
& \textbf{0.091} \downimp{42.0}
& \textbf{0.108} \downimp{57.1}
& \textbf{0.786} \downimp{55.9}
& \textbf{0.193} \gain{0.290}
& \textbf{0.323} \upimp{35.1}
& \textbf{48.2\%} \\
\midrule
\multirow{5}{*}{\gpt}
& Direct
& 1.738 & 0.197 & 0.277 & 2.338 & -0.101 & 0.253 & --- \\
& Few-shot
& 1.272 \downimp{26.8}
& 0.162 \downimp{17.8}
& 0.175 \downimp{36.8}
& 1.698 \downimp{27.4}
& 0.124 \gain{0.225}
& 0.285 \upimp{12.6}
& 27.2\% \\
& ToT
& 1.493 \downimp{14.1}
& 0.223 \textcolor{red}{($\uparrow$13.4\%)}
& 0.210 \downimp{24.3}
& 1.821 \downimp{22.1}
& 0.088 \gain{0.189}
& 0.320 \upimp{26.5}
& 11.8\% \\
& RoT
& 1.468 \downimp{15.5}
& 0.172 \downimp{12.6}
& 0.183 \downimp{34.1}
& 1.607 \downimp{31.3}
& -0.100 \gain{0.001}
& 0.241 \textcolor{red}{($\downarrow$4.7\%)}
& 23.4\% \\
& \textbf{Ours}
& \textbf{0.887} \downimp{49.0}
& \textbf{0.154} \downimp{21.8}
& \textbf{0.126} \downimp{54.5}
& \textbf{0.828} \downimp{64.6}
& \textbf{0.190} \gain{0.291}
& \textbf{0.327} \upimp{29.2}
& \textbf{47.5\%} \\
\midrule
\multirow{5}{*}{\llama}
& Direct
& 1.395 & 0.243 & 0.434 & 2.657 & -0.043 & 0.261 & --- \\
& Few-shot
& 1.066 \downimp{23.6}
& 0.142 \downimp{41.6}
& 0.197 \downimp{54.6}
& 1.815 \downimp{31.7}
& 0.124 \gain{0.167}
& 0.303 \upimp{16.1}
& 37.9\% \\
& ToT
& 1.746 \textcolor{red}{($\uparrow$25.2\%)}
& 0.289 \textcolor{red}{($\uparrow$18.7\%)}
& 0.362 \downimp{16.5}
& 2.809 \textcolor{red}{($\uparrow$5.7\%)}
& 0.032 \gain{0.075}
& 0.235 \textcolor{red}{($\downarrow$10.0\%)}
& \textcolor{red}{-8.3\%} \\
& RoT
& 2.138 \textcolor{red}{($\uparrow$53.2\%)}
& 0.344 \textcolor{red}{($\uparrow$41.4\%)}
& 0.570 \textcolor{red}{($\uparrow$31.3\%)}
& 3.878 \textcolor{red}{($\uparrow$45.9\%)}
& -0.088 \textcolor{red}{(-0.045)}
& 0.207 \textcolor{red}{($\downarrow$20.7\%)}
& \textcolor{red}{-43.0\%} \\
& \textbf{Ours}
& \textbf{0.916} \downimp{34.3}
& \textbf{0.133} \downimp{45.3}
& \textbf{0.137} \downimp{68.4}
& \textbf{1.588} \downimp{40.2}
& \textbf{0.149} \gain{0.192}
& \textbf{0.313} \upimp{19.9}
& \textbf{47.1\%} \\
\midrule
\multicolumn{2}{l}{\textbf{Ours (Avg.)}}
& \textbf{0.766}
& \textbf{0.107}
& \textbf{0.102}
& \textbf{0.906}
& \textbf{0.189}
& \textbf{0.330}
& \textbf{54.4\%} \\
\bottomrule
\end{tabular}
\caption{
\textbf{Main Results}
Each cell is the distance between a method's outputs and real platform answers. \textbf{Avg.\ Shrink} is the mean relative reduction over the four distributional metrics against Direct. Parenthesized values are relative changes over Direct (for $\rho$: absolute point gain), with \textcolor{green!50!black}{green} marking improvement and \textcolor{red}{red} marking degradation. The last row averages OMRA over the four base models.
}
\label{tab:gap_and_OMRA}
\end{table*}

\paragraph{Over-Explained Span Identification}
We define \emph{over-explained spans} as surface spans in the final
answer that make its reasoning unnecessarily explicit, repetitive, or
formulaic.
They include claim--evidence bridges, justification-heavy sentences,
causal explanations, enumerated steps, and explicit transitions.
These spans are not hidden reasoning traces.
OMRA reuses the ontology extractor rather than training a separate span
classifier: each unit
$u \in \mathcal{O}^{L}(a)\cup\mathcal{O}^{E}(a)$ is aligned with its
source span, and units whose relation or pattern type belongs to
$\mathcal{T}_{\mathrm{over}}$ form $\mathcal{S}(a)$.

\paragraph{Ontology-Guided Masking}
OMRA replaces spans in $\mathcal{S}(a)$ with mask tokens while preserving
the main stance and surrounding context.

\paragraph{Reconstruction with Preference Constraints}
The reconstructor $\mathcal{R}$ fills the masks under four constraints:
(i) stance preservation,
(ii) factual consistency with non-masked content,
(iii) discourse coherence, and
(iv) affective or expressive salience only when contextually supported.
The goal is not to recover the original spans, but to express the same
answer with less over-explanation.

\paragraph{Preference-Aware Revision}
A verification pass $\mathcal{V}$ checks whether $\hat{a}$ still contains
over-explanation associated with logic overbinding.
If so, $\mathcal{V}$ applies a lightweight revision; otherwise,
$a'=\hat{a}$.

\section{Experiments}
\label{sec:experiments}

\subsection{Experimental Setup}
\label{sec:exp_setup}

We evaluate on a controlled benchmark of 3{,}600 real platform answers sampled for matched within-question comparison. Each selected question provides one representative answer at each of the four within-question engagement levels (Appendix~\ref{app:benchmark_sampling}). For each question--level pair, we construct three generated regimes: \textbf{Direct (Synthetic)}, generated directly under the target engagement level; \textbf{Few-shot}, additionally conditioned on a level-matched real reference; and \textbf{OMRA-corrected}, revised by Algorithm~\ref{alg:OMRA}. We instantiate the pipeline with four LLM families---DeepSeek, Claude, GPT, and Llama---and compare against two reasoning-augmented baselines, Tree-of-Thought (ToT)~\cite{yao2023tree} and Ripple-of-Thought (RoT)~\cite{lei2025godbench}.

We use DeepSeek-V3.2 as the fixed ontology extractor across all four model families.
To validate its reliability, we compare its scores with human consensus on a randomly
selected 100-question subset, obtaining Spearman correlations of 0.932, 0.873,
and 0.935 for Logic, Affect, and Expression, respectively (0.953 overall).
Further validation details are provided in Appendix~\ref{app:scoring_reliability}.

\begin{figure*}[t]
    \centering
    \includegraphics[width=0.98\textwidth]{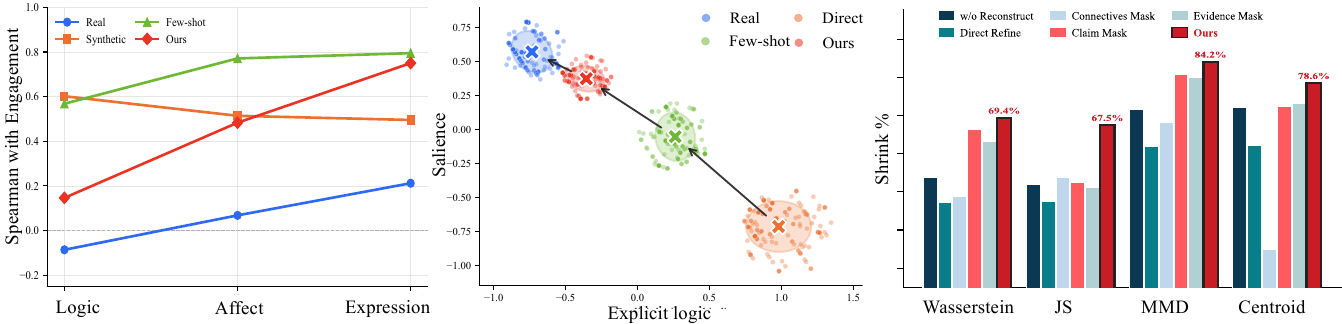}
    \caption{
    \textbf{Diagnosing and reducing the AI preference--user engagement gap.}
    \textbf{(a)} Dimension-wise Spearman $\rho$ between ontology scores and within-question engagement levels across answer regimes.
    \textbf{(b)} Joint preference geometry in a diagnostic space of logic overbinding and affective--expressive salience.
    \textbf{(c)} Ablation over masking and revision variants in OMRA on DeepSeek.
    }
    \label{fig:main_results}
\end{figure*}

\begin{table}[t]
\centering
\small
\setlength{\tabcolsep}{4.5pt}
\renewcommand{\arraystretch}{1.08}
\begin{tabular}{lcccc}
\toprule
\textbf{Method}
& \textbf{Avg. $\rho$}
& \textbf{Med. $\rho$}
& \textbf{Top-1}
& \textbf{Pearson} \\
& & & \textbf{Acc.} & $\boldsymbol{r}$ \\
\midrule
Random baseline
& 0.000 & 0.000 & 0.250 & 0.000 \\
Text embedding
& 0.214 & 0.400 & 0.324 & 0.309 \\
\midrule
LLM judge
& & & & \\
\cmidrule(lr){1-5}
~~ DeepSeek
& -0.080 & -0.200 & 0.141 & -0.050 \\
~~ ChatGPT
& -0.127 & -0.258 & 0.140 & -0.075 \\
~~ Llama
& -0.004 & 0.258 & 0.082 & -0.014 \\
~~ Claude
& 0.182 & 0.258 & 0.164 & 0.121 \\
\midrule
\textbf{Ontology features}
& \textbf{0.359}
& \textbf{0.400}
& \textbf{0.435}
& \textbf{0.465} \\
\bottomrule
\end{tabular}
\caption{
Validation of ontological preference measurement on real-user answers.
All ranking metrics are computed within questions.
}
\label{tab:real_preference_prediction}
\end{table}

\subsection{Evaluation Results}
\label{sec:eval_results}

\paragraph{Ontology features capture signals associated with real user engagement better than dense embeddings and LLM judges.}
Before evaluating OMRA, we verify that the ontology vector $\mathbf{s}(a)$ carries engagement-relevant signals in real platform answers. Table~\ref{tab:real_preference_prediction} compares a lightweight predictor over $\mathbf{s}(a)$ with dense text embeddings and four LLM-as-a-judge baselines on within-question ranking. Ontology features outperform all baselines, reaching Avg.\ $\rho=0.359$ and Top-1 $=0.435$, while three of four LLM judges fall below the dense-embedding baseline. These results suggest that the ontology representation captures real user engagement signals that black-box LLM judges do not reliably recover.

\paragraph{OMRA reduces the AI preference--user engagement gap across all four LLM families.}
Table~\ref{tab:gap_and_OMRA} compares OMRA with Direct, Few-shot, ToT, and RoT across marginal, geometric, and transfer metrics. OMRA reduces the measured gap by an average of \textbf{54.4\%} and is the only method that improves all three views on every base model. Few-shot is the strongest baseline but remains weaker on geometry and transfer, suggesting that reference imitation does not fully recover the patterns associated with real user engagement. ToT and RoT do not consistently reduce the gap across model families, suggesting that reasoning augmentation alone does not recover these patterns. OMRA performs consistently across all four model families, indicating that its improvement is not specific to a single generator.

\paragraph{Additional controls rule out simple alternative explanations.}
We conduct additional diagnostic experiments on DeepSeek (Appendix~\ref{app:additional_controls}). Length control improves both Direct and Few-shot generation, but does not account for OMRA's gains. Platform-aware prompting provides modest improvements, while sentence-mood constraints have only minor effects. A separate preservation evaluation further shows that OMRA largely preserves stance, factual consistency, answer relevance, and usefulness.

\subsection{Ablation Study}
\label{sec:ablation}

\paragraph{Joint span identification is more effective than surface-marker or single-layer masking.}
We ablate OMRA with five variants in Figure~\ref{fig:main_results}(c): \textbf{w/o Reconstruct} leaves masked spans empty; \textbf{Direct Refine} rewrites the answer without masking; \textbf{Connectors Mask} masks only surface discourse markers; and \textbf{Claim Mask} / \textbf{Evidence Mask} mask only one component type. Full OMRA performs best across all gap metrics. Direct Refine and Connectors Mask yield the weakest improvements, showing that generic rewriting or surface-marker deletion is insufficient. Claim Mask and Evidence Mask achieve intermediate results, suggesting that over-explanation is better captured through joint spans involving claims, evidence, and their relations.
\begin{table}[t]
\centering
\small
\setlength{\tabcolsep}{6pt}
\renewcommand{\arraystretch}{1.08}
\begin{tabular}{lcc}
\toprule
\textbf{Training $\to$ Test Regime}
& \textbf{Spearman} $\boldsymbol{\rho}$
& \textbf{Top-1} \\
\midrule
Synthetic $\to$ Synthetic & 0.947 & 0.895 \\
Real $\to$ Synthetic      & 0.759 & 0.800 \\
Synthetic $\to$ Real      & $-$0.073 & 0.254 \\
\bottomrule
\end{tabular}
\caption{
\textbf{Cross-regime transfer of ontological preference predictors.}
A predictor is trained on the source regime and evaluated on the target regime; metrics are averaged over four base LLM families.
}
\label{tab:transfer_asymmetry}
\end{table}

\section{Analysis}
\label{sec:analysis}

\subsection{AI Preference--User Engagement Gap}
\label{sec:findings}

We examine cross-regime transfer, dimension-wise associations,
and joint preference geometry (\S\ref{sec:discovery}).
These comparisons characterize associations with observed
engagement rather than causal effects of textual features.

\begin{table*}[t]
\centering
\scriptsize
\setlength{\tabcolsep}{5pt}
\renewcommand{\arraystretch}{1.10}
\begin{tabular}{lccccc}
\toprule
\textbf{Pairing}
& $\boldsymbol{N}$
& \textbf{Human-like Win Rate}
& \textbf{Preferred Win Rate}
& $\boldsymbol{\kappa}$
& \textbf{Reversal Rate} \\
\midrule
Direct vs Real
& 134
& 6.7\% [3.6, 12.3]
& 60.4\% [52.0, 68.3]
& 0.672
& 56.7\% [48.3, 64.8] \\

Few-shot vs Real
& 134
& 25.4\% [18.8, 33.4]
& 63.4\% [55.0, 71.1]
& 0.760
& 45.5\% [37.3, 54.0] \\

\textbf{OMRA vs Real}
& \textbf{132}
& \textbf{31.1\% [23.8, 39.4]}
& \textbf{67.4\% [59.0, 74.8]}
& \textbf{0.727}
& \textbf{43.9\% [35.8, 52.5]} \\
\bottomrule
\end{tabular}
\caption{
\textbf{Human evaluation against matched real platform answers.}
Results are based on 100 human annotators.
Win rates and reversal rates include 95\% confidence intervals;
$\kappa$ denotes inter-annotator agreement.
}
\label{tab:human_eval}
\end{table*}

\begin{findingbox}{Finding 1: AI preference does not transfer to real user engagement.}
A predictor trained on Synthetic answers cannot reliably rank real platform answers, although a predictor trained on real answers transfers reasonably well to Synthetic answers.
\end{findingbox}

\noindent
Averaged over four base models (Table~\ref{tab:transfer_asymmetry}), Synthetic$\to$Synthetic reaches $\rho=0.947$ and Real$\to$Synthetic reaches $\rho=0.759$, whereas Synthetic$\to$Real drops to $\rho=-0.073$. This asymmetry shows that internal consistency in AI preference does not imply transfer to real user engagement.
Additional predictor robustness and SHAP analyses confirm that the transfer failure persists across learned estimators and is accompanied by greater reliance on Logic features in Synthetic-trained predictors (Appendix~\ref{app:predictor_robustness}).

\begin{findingbox}{Finding 2: AI preference overweights explicit logic relative to real user engagement.}
Explicit logical structure is positively associated with target engagement in Synthetic answers, whereas real user engagement is more strongly associated with affective and expressive salience.
\end{findingbox}

\noindent
Figure~\ref{fig:main_results}(a) reveals \textbf{logic overbinding}:
Logic is strongly associated with target engagement in Synthetic answers
but weakly negatively associated with real engagement, where Affect and
Expression show stronger positive associations.
This pattern persists across question-property strata
(Appendix~\ref{app:question_stratification}), while ToT and RoT
fail to consistently reduce the gap across model families.

\begin{findingbox}{Finding 3: AI-generated answers occupy a displaced preference geometry.}
Real and AI-generated answers combine logic, affect, and expression differently, placing them in distinct regions of the joint preference space.
\end{findingbox}

\noindent
Figure~\ref{fig:main_results}(b) shows that Direct answers occupy a more logic-heavy and lower-salience region than real platform answers. Few-shot moves partially toward the real-answer distribution, while OMRA moves substantially closer. The MMD and centroid results in Table~\ref{tab:gap_and_OMRA} provide corresponding quantitative evidence.

\subsection{Human Evaluation}
\label{sec:human_eval}

Findings 1--3 characterize the gap in the ontological preference space. We next examine whether reducing this gap requires making answers appear human-written.

\paragraph{Setup.}
We compare each method's answers with matched real platform answers. Each item is evaluated by all 100 annotators, who answer two questions: (i) which answer appears more human-written, and (ii) which answer they would prefer to like or endorse. Full instructions, interface details, and the reversal-rate definition are provided in Appendix~\ref{app:human_eval}.

\begin{findingbox}{Finding 4: Pairwise preference judgments can diverge from perceived human-likeness.}
Annotators may prefer an AI-generated answer even when they judge the matched real answer as more likely to be human-written.
\end{findingbox}

\noindent
Direct answers are judged less human-like in 93.3\% of
comparisons, yet win 60.4\% of pairwise preference judgments.
OMRA achieves the highest preferred win rate at 67.4\%,
while its perceived human-likeness win rate is 31.1\%.
Since the confidence intervals for OMRA and Few-shot overlap,
we treat this result as supportive rather than statistically
decisive evidence.
The observed reversals indicate that perceived human-likeness
and pairwise preference are not identical judgments.
However, asking both questions in the same interface may
introduce priming effects, which these results cannot rule out.

\section{Conclusion}
\label{sec:conclusion}

We introduced \textbf{Ontological Preference Measurement} to compare AI preference and real user engagement in an interpretable space. Our analysis reveals a gap across transfer, marginal, and geometric comparisons, characterized by \textbf{logic overbinding}: AI preference overweights explicit logical presentation. To address this, we proposed \textbf{Ontology-Masked Reasoning Autoencoding (OMRA)}, which reconstructs over-explained spans and reduces the measured gap by \textbf{54.4\%} on average across four LLM families. In human evaluation with 100 annotators, OMRA wins \textbf{67.4\%} of pairwise preferences against matched real answers, although these answers are more often judged human-written. This suggests that pairwise preference is not reducible to perceived human-likeness.

\section*{Limitations}

\textbf{(1) Dependence on ontology extraction quality.}
Our framework relies on LLM-based ontology extraction.
Although the extractor shows strong agreement with human annotations, errors may still occur for implicit affect, sarcasm, culturally specific rhetoric, and highly context-dependent expressions.

\noindent
\textbf{(2) Inference-time correction does not change the model's internal preference.}
OMRA revises generated answers at inference time, but does not directly modify the model's underlying preference representation or decoding behavior.
Future work may explore more durable model-level alignment methods.

\noindent
\textbf{(3) Offline evaluation, data constraints, and dual-use risks.}
OMRA is evaluated offline and pairwise preference does not fully reproduce real platform dynamics.
We therefore use OMRA as a controlled intervention rather than a tool for maximizing engagement.
Our corpus contains publicly visible answer-level data without user profiling; released resources will remove user identifiers and follow platform-specific redistribution constraints.
Live deployment would require additional ethical safeguards against engagement manipulation and large-scale content generation.
\bibliography{main}

@article{gruber1993translation,
  title={A translation approach to portable ontology specifications},
  author={Gruber, Thomas R},
  journal={Knowledge acquisition},
  volume={5},
  number={2},
  pages={199--220},
  year={1993},
  publisher={Elsevier}
}

@misc{noy2001ontology,
  title={Ontology development 101: A guide to creating your first ontology},
  author={Noy, Natalya F and McGuinness, Deborah L and others},
  year={2001},
  publisher={Stanford knowledge systems laboratory technical report KSL-01-05 and~…}
}

@article{lippi2016argumentation,
  title={Argumentation mining: State of the art and emerging trends},
  author={Lippi, Marco and Torroni, Paolo},
  journal={ACM Transactions on Internet Technology (TOIT)},
  volume={16},
  number={2},
  pages={1--25},
  year={2016},
  publisher={ACM New York, NY, USA}
}

@article{lawrence2019argument,
  title={Argument mining: A survey},
  author={Lawrence, John and Reed, Chris},
  journal={Computational linguistics},
  volume={45},
  number={4},
  pages={765--818},
  year={2019}
}

@article{stab2017parsing,
  title={Parsing argumentation structures in persuasive essays},
  author={Stab, Christian and Gurevych, Iryna},
  journal={Computational Linguistics},
  volume={43},
  number={3},
  pages={619--659},
  year={2017}
}

@inproceedings{lei2025godbench,
  title={GODBench: A Benchmark for Multimodal Large Language Models in Video Comment Art},
  author={Lei, Yiming and Zhang, Chenkai and Liu, Zeming and Leng, Haitao and Liu, Shaoguo and Gao, Tingting and Liu, Qingjie and Wang, Yunhong},
  booktitle={Proceedings of the 63rd Annual Meeting of the Association for Computational Linguistics (Volume 1: Long Papers)},
  pages={11884--11952},
  year={2025}
}

@inproceedings{ye2025mvp,
  title={Mvp: Winning solution to smp challenge 2025 video track},
  author={Ye, Liliang and Zhang, Yunyao and Wu, Yafeng and Chen, Yi-Ping Phoebe and Yu, Junqing and Yang, Wei and Song, Zikai},
  booktitle={Proceedings of the 33rd ACM International Conference on Multimedia},
  pages={14079--14085},
  year={2025}
}

@article{wang2026seeing,
  title={Seeing Further and Wider: Joint Spatio-Temporal Enlargement for Micro-Video Popularity Prediction},
  author={Wang, Dali and Zhang, Yunyao and Yu, Junqing and Chen, Yi-Ping Phoebe and Xu, Chen and Song, Zikai},
  journal={arXiv preprint arXiv:2604.20311},
  year={2026}
}

@article{wu2026hotcomment,
  title={HotComment: A Benchmark for Evaluating Popularity of Online Comments},
  author={Wu, Yafeng and Zhang, Yunyao and Ye, Liliang and Zeng, Guiyi and Yu, Junqing and Xu, Chen and Song, Zikai},
  journal={arXiv preprint arXiv:2604.25614},
  year={2026}
}

@article{zhang2026intervensim,
  title={IntervenSim: Intervention-Aware Social Network Simulation for Opinion Dynamics},
  author={Zhang, Yunyao and Ying, Zuocheng and Zhang, Xinglang and Yu, Junqing and Fang, Peng and Chen, Xu and Yang, Wei and Song, Zikai},
  journal={arXiv preprint arXiv:2604.06600},
  year={2026}
}

@article{zhang2026coupling,
  title={Coupling Macro Dynamics and Micro States for Long-Horizon Social Simulation},
  author={Zhang, Yunyao and Ai, Yihao and Ying, Zuocheng and Mi, Qirui and Yu, Junqing and Yang, Wei and Song, Zikai},
  journal={arXiv preprint arXiv:2604.05516},
  year={2026}
}

@article{deng2020contributes,
  title={Who contributes what? Scrutinizing the activity data of 4.2 million Zhihu users via immersion scores},
  author={Deng, Shengli and Jiang, Yuting and Li, Hongxiu and Liu, Yong},
  journal={Information Processing \& Management},
  volume={57},
  number={5},
  pages={102274},
  year={2020},
  publisher={Elsevier}
}

@inproceedings{rameez2022viralbert,
  title={ViralBERT: A user focused BERT-based approach to virality prediction},
  author={Rameez, Rikaz and Rahmani, Hossein A and Yilmaz, Emine},
  booktitle={Adjunct Proceedings of the 30th ACM Conference on User Modeling, Adaptation and Personalization},
  pages={85--89},
  year={2022}
}

@article{li2024crowdsourced,
  title={From crowdsourced data to high-quality benchmarks: Arena-hard and benchbuilder pipeline},
  author={Li, Tianle and Chiang, Wei-Lin and Frick, Evan and Dunlap, Lisa and Wu, Tianhao and Zhu, Banghua and Gonzalez, Joseph E and Stoica, Ion},
  journal={arXiv preprint arXiv:2406.11939},
  year={2024}
}

@article{christiano2017deep,
  title={Deep reinforcement learning from human preferences},
  author={Christiano, Paul F and Leike, Jan and Brown, Tom and Martic, Miljan and Legg, Shane and Amodei, Dario},
  journal={Advances in neural information processing systems},
  volume={30},
  year={2017}
}

@article{ouyang2022training,
  title={Training language models to follow instructions with human feedback},
  author={Ouyang, Long and Wu, Jeffrey and Jiang, Xu and Almeida, Diogo and Wainwright, Carroll and Mishkin, Pamela and Zhang, Chong and Agarwal, Sandhini and Slama, Katarina and Ray, Alex and others},
  journal={Advances in neural information processing systems},
  volume={35},
  pages={27730--27744},
  year={2022}
}

@inproceedings{lambert2025rewardbench,
  title={Rewardbench: Evaluating reward models for language modeling},
  author={Lambert, Nathan and Pyatkin, Valentina and Morrison, Jacob and Miranda, LJ and Lin, Bill Yuchen and Chandu, Khyathi and Dziri, Nouha and Kumar, Sachin and Zick, Tom and Choi, Yejin and others},
  booktitle={Findings of the Association for Computational Linguistics: NAACL 2025},
  pages={1755--1797},
  year={2025}
}

@article{hu2024explaining,
  title={Explaining length bias in llm-based preference evaluations},
  author={Hu, Zhengyu and Song, Linxin and Zhang, Jieyu and Xiao, Zheyuan and Wang, Tianfu and Chen, Zhengyu and Yuan, Nicholas Jing and Lian, Jianxun and Ding, Kaize and Xiong, Hui},
  journal={arXiv preprint arXiv:2407.01085},
  year={2024}
}

@inproceedings{ye2025justice,
  title={Justice or prejudice? quantifying biases in llm-as-a-judge},
  author={Ye, Jiayi and Wang, Yanbo and Huang, Yue and Chen, Dongping and Zhang, Qihui and Moniz, Nuno and Gao, Tian and Geyer, Werner and Huang, Chao and Chen, Pin-Yu and others},
  booktitle={International Conference on Learning Representations},
  volume={2025},
  pages={102351--102390},
  year={2025}
}

@article{saito2023verbosity,
  title={Verbosity bias in preference labeling by large language models},
  author={Saito, Keita and Wachi, Akifumi and Wataoka, Koki and Akimoto, Youhei},
  journal={arXiv preprint arXiv:2310.10076},
  year={2023}
}

@inproceedings{wang2025eliminating,
  title={Eliminating position bias of language models: A mechanistic approach},
  author={Wang, Ziqi and Zhang, Hanlin and Li, Xiner and Huang, Kuan-Hao and Han, Chi and Ji, Shuiwang and Kakade, Sham and Peng, Hao and Ji, Heng},
  booktitle={International Conference on Learning Representations},
  volume={2025},
  pages={91212--91239},
  year={2025}
}

@article{zhang2023survey,
  title={A survey of controllable text generation using transformer-based pre-trained language models},
  author={Zhang, Hanqing and Song, Haolin and Li, Shaoyu and Zhou, Ming and Song, Dawei},
  journal={ACM Computing Surveys},
  volume={56},
  number={3},
  pages={1--37},
  year={2023},
  publisher={ACM New York, NY}
}

@inproceedings{dong2025self,
  title={Self-boosting large language models with synthetic preference data},
  author={Dong, Qingxiu and Dong, Li and Zhang, Xingxing and Sui, Zhifang and Wei, Furu},
  booktitle={International Conference on Learning Representations},
  volume={2025},
  pages={65440--65463},
  year={2025}
}

@article{yao2023tree,
  title={Tree of thoughts: Deliberate problem solving with large language models},
  author={Yao, Shunyu and Yu, Dian and Zhao, Jeffrey and Shafran, Izhak and Griffiths, Tom and Cao, Yuan and Narasimhan, Karthik},
  journal={Advances in neural information processing systems},
  volume={36},
  pages={11809--11822},
  year={2023}
}

@article{rafailov2023direct,
  title={Direct preference optimization: Your language model is secretly a reward model},
  author={Rafailov, Rafael and Sharma, Archit and Mitchell, Eric and Manning, Christopher D and Ermon, Stefano and Finn, Chelsea},
  journal={Advances in neural information processing systems},
  volume={36},
  pages={53728--53741},
  year={2023}
}

@inproceedings{chakrabarty2019ampersand,
  title={AMPERSAND: Argument mining for PERSuAsive oNline discussions},
  author={Chakrabarty, Tuhin and Hidey, Christopher and Muresan, Smaranda and Mckeown, Kathleen and Hwang, Alyssa},
  booktitle={Proceedings of the 2019 Conference on Empirical Methods in Natural Language Processing and the 9th International Joint Conference on Natural Language Processing (EMNLP-IJCNLP)},
  pages={2933--2943},
  year={2019}
}

@article{wataoka2024self,
  title={Self-preference bias in llm-as-a-judge},
  author={Wataoka, Koki and Takahashi, Tsubasa and Ri, Ryokan},
  journal={arXiv preprint arXiv:2410.21819},
  year={2024}
}

@article{zhang2023human,
  title={Human favoritism, not AI aversion: People’s perceptions (and bias) toward generative AI, human experts, and human--GAI collaboration in persuasive content generation},
  author={Zhang, Yunhao and Gosline, Ren{\'e}e},
  journal={Judgment and Decision Making},
  volume={18},
  pages={e41},
  year={2023},
  publisher={Cambridge University Press}
}

@article{rallapalli2026interpretable,
  title={Interpretable Stylistic Variation in Human and LLM Writing Across Genres, Models, and Decoding Strategies},
  author={Rallapalli, Swati and Gallagher, Shannon and Yurko, Ronald and Brooks, Tyler and Loughin, Chuck and Sezgin, Michele and Turri, Violet},
  journal={arXiv preprint arXiv:2604.14111},
  year={2026}
}

@article{rodrigues2026linguistic,
  title={A linguistic comparison between human-and AI-generated content},
  author={Rodrigues, Fl{\'a}via A and Sturm, Niclas F and Pinheiro, Fl{\'a}vio L},
  journal={Iscience},
  volume={29},
  number={3},
  year={2026},
  publisher={Elsevier}
}

@article{li2025preference,
  title={Preference leakage: A contamination problem in llm-as-a-judge},
  author={Li, Dawei and Sun, Renliang and Huang, Yue and Zhong, Ming and Jiang, Bohan and Han, Jiawei and Zhang, Xiangliang and Wang, Wei and Liu, Huan},
  journal={arXiv preprint arXiv:2502.01534},
  year={2025}
}

@article{william1988rhetorical,
  title={Rhetorical structure theory: Towards a functional theory of text organization},
  author={William, Mann and Thompson, Sandra},
  journal={Text},
  volume={8},
  number={3},
  pages={243--281},
  year={1988}
}

@inproceedings{feng2011classifying,
  title={Classifying arguments by scheme},
  author={Feng, Vanessa Wei and Hirst, Graeme},
  booktitle={Proceedings of the 49th annual meeting of the association for computational linguistics: Human language technologies},
  pages={987--996},
  year={2011}
}

@inproceedings{tan2016winning,
  title={Winning arguments: Interaction dynamics and persuasion strategies in good-faith online discussions},
  author={Tan, Chenhao and Niculae, Vlad and Danescu-Niculescu-Mizil, Cristian and Lee, Lillian},
  booktitle={Proceedings of the 25th international conference on world wide web},
  pages={613--624},
  year={2016}
}

@inproceedings{he2022masked,
  title={Masked autoencoders are scalable vision learners},
  author={He, Kaiming and Chen, Xinlei and Xie, Saining and Li, Yanghao and Doll{\'a}r, Piotr and Girshick, Ross},
  booktitle={Proceedings of the IEEE/CVF conference on computer vision and pattern recognition},
  pages={16000--16009},
  year={2022}
}

@inproceedings{liu2025learning,
  title={Learning to Substitute Words with Model-based Score Ranking},
  author={Liu, Hongye and Henao, Ricardo},
  booktitle={Proceedings of the 2025 Conference of the Nations of the Americas Chapter of the Association for Computational Linguistics: Human Language Technologies (Volume 1: Long Papers)},
  pages={11551--11565},
  year={2025}
}

@article{liu2026learning,
  title={Learning to Control Summaries with Score Ranking},
  author={Liu, Hongye and Ding, Liang and Henao, Ricardo},
  journal={arXiv preprint arXiv:2604.17197},
  year={2026}
}

@article{liu2026calibrating,
  title={Calibrating Model-Based Evaluation Metrics for Summarization},
  author={Liu, Hongye and Brahma, Dhanajit and Henao, Ricardo},
  journal={arXiv preprint arXiv:2604.17200},
  year={2026}
}

@article{zhou2026exoactor,
  title={ExoActor: Exocentric Video Generation as Generalizable Interactive Humanoid Control},
  author={Zhou, Yanghao and Ma, Jingyu and Peng, Yibo and Sun, Zhenguo and Bai, Yu and Karlsson, B{\"o}rje F},
  journal={arXiv preprint arXiv:2604.27711},
  year={2026}
}

@inproceedings{jin2026simtoken,
  title={Simtoken: A simple baseline for referring audio-visual segmentation},
  author={Jin, Dian and Zhou, Yanghao and Zhou, Jinxing and Ma, Jiaqi and Guo, Ruohao and Guo, Dan},
  booktitle={ICASSP 2026-2026 IEEE International Conference on Acoustics, Speech and Signal Processing (ICASSP)},
  pages={22702--22706},
  year={2026},
  organization={IEEE}
}

@inproceedings{zhang-etal-2026-logical,
    title = "Logical Phase Transitions: Understanding Collapse in {LLM} Logical Reasoning",
    author = "Zhang, Xinglang  and
      Zhang, Yunyao  and
      Chen, ZeLiang  and
      Yu, Junqing  and
      Yang, Wei  and
      Song, Zikai",
    editor = "Liakata, Maria  and
      Moreira, Viviane P.  and
      Zhang, Jiajun  and
      Jurgens, David",
    booktitle = "Proceedings of the 64th Annual Meeting of the {A}ssociation for {C}omputational {L}inguistics (Volume 1: Long Papers)",
    month = jul,
    year = "2026",
    address = "San Diego, California, United States",
    publisher = "Association for Computational Linguistics",
    url = "https://aclanthology.org/2026.acl-long.858/",
    doi = "10.18653/v1/2026.acl-long.858",
    pages = "18836--18860",
    ISBN = "979-8-89176-390-6"
}

@inproceedings{zhang-etal-2026-semantic,
    title = "Semantic-Aware Logical Reasoning via a Semiotic Framework",
    author = "Zhang, Yunyao  and
      Zhang, Xinglang  and
      Sheng, Junxi  and
      Li, Wenbing  and
      Yu, Junqing  and
      Chen, Yi-Ping Phoebe  and
      Yang, Wei  and
      Song, Zikai",
    editor = "Liakata, Maria  and
      Moreira, Viviane P.  and
      Zhang, Jiajun  and
      Jurgens, David",
    booktitle = "Proceedings of the 64th Annual Meeting of the {A}ssociation for {C}omputational {L}inguistics (Volume 1: Long Papers)",
    month = jul,
    year = "2026",
    address = "San Diego, California, United States",
    publisher = "Association for Computational Linguistics",
    url = "https://aclanthology.org/2026.acl-long.835/",
    doi = "10.18653/v1/2026.acl-long.835",
    pages = "18349--18374",
    ISBN = "979-8-89176-390-6"
}

@article{zheng2026surfacing,
  title={Surfacing the Unsaid: CUE-Bench for Affective Stance in Chinese Discourse},
  author={Zheng, Zhenyan and Zhang, Yunyao and Sheng, Junxi and Yu, Junqing and Song, Zikai},
  journal={arXiv preprint arXiv:2608.10810},
  year={2026}
}

@misc{song2026socialintelligence,
  title        = {Social Intelligence Modeling: A Comprehensive Survey from Social Perception to Social Simulation},
  author       = {Song, Zikai and Li, Xiajie and Zhang, Yunyao and Zhang, Xinglang and Yang, Wei and Yu, Junqing},
  howpublished = {ResearchGate preprint},
  year         = {2026},
  note         = {Preprint available on ResearchGate},
  doi          = {10.13140/RG.2.2.21157.87528},
  url          = {https://www.researchgate.net/publication/411123257_Social_Intelligence_Modeling_A_Comprehensive_Survey_from_Social_Perception_to_Social_Simulation}
}

@article{chen2026show,
  title={Show, Don't Tell: Morphing Latent Reasoning into Image Generation},
  author={Chen, Harold Haodong and Yin, Xinxiang and Shu, Wen-Jie and Zhang, Hongfei and Zhang, Zixin and Liao, Chenfei and Guo, Litao and Chen, Qifeng and Chen, Ying-Cong},
  journal={arXiv preprint arXiv:2602.02227},
  year={2026}
}

@article{cui2026earthverse,
  title={EarthVerse: Benchmarking Scientific Agents Across Dynamic Earth Systems and Natural Hazards},
  author={Cui, Zhiqing and Yin, Xinxiang and Tang, Yihong and Zhang, Xinglang and Hu, Yuanzhe and Zhong, Siru and Tang, Weidong and Liang, Yuxuan and Li, Weijia and Jin, Ming and others},
  journal={arXiv preprint arXiv:2608.23525},
  year={2026}
}

@inproceedings{cui2026augur,
  title={Augur: Modeling covariate causal associations in time series via large language models},
  author={Cui, Zhiqing and Wang, Binwu and Liu, Qingxiang and Wang, Yeqiang and Zhou, Zhengyang and Liang, Yuxuan and Wang, Yang},
  booktitle={Proceedings of the 64th Annual Meeting of the Association for Computational Linguistics (Volume 1: Long Papers)},
  pages={764--787},
  year={2026}
}

@inproceedings{ma2025causal,
  title={Causal Learning Meet Covariates: Empowering Lightweight and Effective Nationwide Air Quality Forecasting.},
  author={Ma, Jiaming and Cui, Zhiqing and Wang, Binwu and Wang, Pengkun and Zhou, Zhengyang and Zhao, Zhe and Wang, Yang},
  booktitle={IJCAI},
  pages={3171--3179},
  year={2025}
}

@inproceedings{li2026lora,
  title={Lora-mixer: Coordinate modular lora experts through serial attention routing},
  author={Li, Wenbing and Song, Zikai and Zhou, Hang and Yu, Junqing and Zhang, Yunyao and Yang, Wei},
  booktitle={International Conference on Learning Representations},
  volume={2026},
  pages={14694--14716},
  year={2026}
}

@article{liu2025deepseek,
  title={Deepseek-v3. 2: Pushing the frontier of open large language models},
  author={Liu, Aixin and Mei, Aoxue and Lin, Bangcai and Xue, Bing and Wang, Bingxuan and Xu, Bingzheng and Wu, Bochao and Zhang, Bowei and Lin, Chaofan and Dong, Chen and others},
  journal={arXiv preprint arXiv:2512.02556},
  year={2025}
}

@misc{touvron2023llama,
      title={LLaMA: Open and Efficient Foundation Language Models}, 
      author={Hugo Touvron and Thibaut Lavril and Gautier Izacard and Xavier Martinet and Marie-Anne Lachaux and Timothée Lacroix and Baptiste Rozière and Naman Goyal and Eric Hambro and Faisal Azhar and Aurelien Rodriguez and Armand Joulin and Edouard Grave and Guillaume Lample},
      year={2023},
      eprint={2302.13971},
      archivePrefix={arXiv},
      primaryClass={cs.CL},
      url={https://arxiv.org/abs/2302.13971}, 
}

@inproceedings{shi2026codehacker,
  title={Codehacker: Automated test case generation for detecting vulnerabilities in competitive programming solutions},
  author={Shi, Jingwei and Yin, Xinxiang and Huang, Jing and Tao, Shengyu and Zhao, Jinman},
  booktitle={Proceedings of the 64th Annual Meeting of the Association for Computational Linguistics (Volume 1: Long Papers)},
  pages={2352--2382},
  year={2026}
}

@article{zhang2026valg,
  title={Valg: An agentic system for ml theory research},
  author={Zhang, Dechen and Tang, Xuan and Yin, Xinxiang and Chen, Xingwu and Qian, Jian and Zou, Difan},
  journal={arXiv preprint arXiv:2608.13060},
  year={2026}
}

@inproceedings{tang2026grouptom,
  title={GroupToM-Bench: Benchmarking Group Theory of Mind and Nonlinear Social Emergence in MLLMs},
  author={Tang, Weidong and Li, Jierui and Hou, Yueling and Mei, Zihan and Zhang, Can and Wan, Xinyan and Liang, Zhiyuan and Zhou, Pengfei and You, Yang and Zhao, Wangbo},
  booktitle={Proceedings of the 64th Annual Meeting of the Association for Computational Linguistics (Volume 1: Long Papers)},
  pages={40007--40031},
  year={2026}
}

@article{tang2026trinity,
  title={The Trinity: Capability-Driven Recursive Self-Improvement for Multimodal Agents},
  author={Tang, Weidong and Wang, Ruiheng and Zhang, Jingxuan and Wang, Hexin and Chen, Haoyuan and Zhang, Zhengfeiyang and Ma, Yixing and Jiang, Zichen and Guo, Ziyu and Li, Haodong and others},
  journal={arXiv preprint arXiv:2610.36077},
  year={2026}
}

@inproceedings{tang2026efficient,
  title={Efficient Video Object Segmentation and Tracking with Recurrent Dynamic Submodel},
  author={Tang, Weidong and Liang, Zhiyuan and Wan, Xinyan and Zhu, Chen and Xu, Zhaopan and Zhou, Pengfei and Song, Yan and You, Yang and Zhao, Wangbo},
  booktitle={Proceedings of the IEEE/CVF Conference on Computer Vision and Pattern Recognition},
  pages={20912--20921},
  year={2026}
}

@inproceedings{chen2025finequest,
  title={Finequest: Adaptive knowledge-assisted sports video understanding via agent-of-thoughts reasoning},
  author={Chen, Haodong and Huang, Haojian and Yin, Xinxiang and Shao, Dian},
  booktitle={Proceedings of the 33rd ACM International Conference on Multimedia},
  pages={2909--2918},
  year={2025}
}

\appendix
\clearpage
\section*{\centering Appendix}

\noindent
The appendix provides supplementary details for related work, dataset construction, ontology design, experimental protocols, human evaluation, case studies, and prompts.

\section*{The Usage of LLM}
In accordance with ACL guidelines, we used large language models solely for writing assistance and language refinement. 

\section{Related Work}
\label{app:related_work}

\subsection{Ontology, Discourse Structure, and Argument Mining}

Ontology-based modeling represents a domain through interpretable concepts, relations, and constraints~\cite{gruber1993translation,noy2001ontology}, and has been widely used in NLP to organize discourse and persuasion phenomena. Rhetorical Structure Theory~\cite{william1988rhetorical} provides a classical account of discourse relations, while later work examines how framing, emotion, and rhetorical choices shape persuasion and user response in online discussion~\cite{feng2011classifying,tan2016winning,zheng2026surfacing}.

A closely related line is argument mining, which extracts claims, premises, evidence, and argumentative relations from natural language~\cite{lippi2016argumentation,lawrence2019argument,stab2017parsing}. Prior work develops corpora and models for component identification, relation classification, and argument quality assessment, focusing primarily on how arguments are structured and supported. Our work uses ontology for a different purpose. Rather than analyzing argument structure alone, we organize online answers along logic~\cite{zhang-etal-2026-semantic,zhang-etal-2026-logical}, affect, and expression, allowing Real, Synthetic, Few-shot, and OMRA-corrected answers to be compared in a shared interpretable preference space.

\subsection{Online Engagement Prediction}

Online popularity and engagement have been studied across social media, micro-video, and community discussion settings using signals such as user authority, temporal dynamics~\cite{cui2026earthverse, cui2026augur,zhang2026valg,chen2025finequest}, propagation networks, content features, and neural representations~\cite{deng2020contributes,rameez2022viralbert,ye2025mvp,wang2026seeing,zhang2026intervensim,zhang2026coupling,lei2025godbench,wu2026hotcomment,tang2026grouptom,tang2026trinity,tang2026efficient}. These approaches generally treat engagement as a scalar prediction target and optimize forecasting~\cite{ma2025causal} performance, but rarely examine how logic, affect, rhetoric, and informativeness are differently associated with engagement.

Our setting differs in both object and goal. We focus on long-form open-platform answers and compare answers only within the same question context. Rather than predicting absolute popularity, we examine which textual qualities vary with real user engagement and whether LLMs associate the same qualities with higher target engagement. This motivates our within-question engagement levels and ontology-based decomposition.

\subsection{AI preference and LLM-as-a-Judge}

Preference data underpins reward modeling, RLHF, and alignment research~\cite{christiano2017deep,ouyang2022training,rafailov2023direct,rallapalli2026interpretable,wataoka2024self,li2026lora}, as well as benchmarks for reward-model evaluation~\cite{lambert2025rewardbench,li2024crowdsourced}. Because human annotation is costly, recent work increasingly relies on AI preference data and LLM-as-a-judge evaluation, in which strong models score or compare outputs as proxies for human annotators~\cite{dong2025self}.

A growing literature documents systematic biases in these signals, including verbosity and length bias~\cite{hu2024explaining,saito2023verbosity}, position and order effects~\cite{wang2025eliminating,li2025preference}, and self-preference toward outputs from the same model family~\cite{ye2025justice}. These biases are usually evaluated against benchmark labels or other model judgments. We instead compare \textbf{AI preference}---the qualities LLMs associate with higher target engagement---with \textbf{real user engagement} observed on online platforms. Our results show that internally consistent AI preference does not necessarily transfer to real platform answers and may place disproportionate weight on explicit logical presentation.

\subsection{Preference-Controlled Generation}

Preference-aligned generation optimizes models toward preferred outputs through reward modeling and policy optimization~\cite{ouyang2022training,rafailov2023direct,liu2025learning,liu2026learning,shi2026codehacker}, while controllable generation steers outputs toward attributes such as style, sentiment, topic, or quality~\cite{zhang2023survey,zhang2023human,zhou2026exoactor,jin2026simtoken,liu2026calibrating}. Common approaches include prompting, demonstrations, latent attribute control, classifier guidance, and reward-based fine-tuning.

These methods generally assume that the target preference or attribute can be specified through a reward signal, prompt, or small set of demonstrations. However, they do not directly identify which answer qualities should be changed when AI preference diverges from real user engagement. OMRA addresses this problem using an explicit ontology: it identifies and masks over-explained spans, then reconstructs the answer under stance-preservation, factual-consistency, coherence, and preference constraints. In this way, OMRA provides a targeted intervention derived from the diagnosed pattern rather than relying on unrestricted style imitation.

\section{Dataset Construction and Benchmark Sampling}
\label{app:dataset}

\subsection{Platforms and Collection}

We construct a large-scale corpus of question--answer pairs from three open online platforms with naturally occurring engagement signals: Zhihu, Quora, and Reddit. Zhihu and Quora provide long-form Q\&A content, while Reddit contributes discussion threads from selected question-answering and explanation-oriented communities.

For each platform, we collect publicly visible question--answer pairs and the engagement metadata available at collection time. The corpus spans 2020--2026 and covers 10 primary domains and 35 sub-domains, as summarized in Figure~\ref{fig:data_overview}. We retain multiple answers to the same question because our analysis relies on within-question comparison. The analysis is conducted at the answer level; we do not construct or use user profiles.

\subsection{Filtering and Quality Control}

We apply the following filters before forming the final corpus:

\begin{itemize}[leftmargin=12pt, topsep=2pt, itemsep=0pt]
    \item \textbf{Length filtering.} We remove extremely short or unusually long answers to exclude trivial replies, link-only responses, and low-quality dumps.
    \item \textbf{Deduplication.} We remove near-duplicate answers within the same question using lexical similarity.
    \item \textbf{Availability filtering.} We discard answers marked as deleted, unavailable, or abnormally inaccessible at collection time, and do not retain deleted content.
    \item \textbf{Minimum-answer filtering.} We keep only questions with at least four surviving answers, which is required for defining four within-question engagement levels $\ell(a)\in\{0,1,2,3\}$.
    \item \textbf{Engagement-anomaly filtering.} We remove questions with extreme engagement concentration or other anomalous feedback patterns.
    \item \textbf{Privacy filtering.} We remove usernames, user identifiers, profile links, and other unnecessary personal metadata from released resources.
\end{itemize}

We preserve the natural distribution of each platform rather than explicitly rebalancing the corpus. Corpus-level statistics and domain distributions are reported in Figure~\ref{fig:data_overview}.

\subsection{Within-Question Engagement Levels}

Absolute vote counts are affected by exposure, platform traffic, author reputation, recommendation algorithms, and time. We therefore do not compare raw votes across questions. Instead, for each question $q$ with answer set $\mathcal{A}_q$, we rank answers by their observed platform engagement signal and map them to four ordinal levels:
\[
\ell(a)=\phi_q(\operatorname{rank}_q(v(a)))\in\{0,1,2,3\}.
\]
This construction compares answers only within the same question context, reducing confounds from question exposure and topic popularity. The resulting label is a relative measure of real user engagement rather than a direct measurement of latent human preference.

\subsection{Controlled Benchmark Sampling}
\label{app:benchmark_sampling}

For the controlled experiments in \S\ref{sec:experiments}, we construct a fixed benchmark of 3{,}600 real platform answers. We select questions with sufficiently many surviving answers and clear engagement separation so that all four within-question engagement levels can be represented.

For each of 900 questions, we retain one real answer at each engagement level. This keeps the question fixed while varying engagement level, reducing confounds from topic, exposure, and question-level popularity. For each question--level pair, we generate a matched Synthetic answer conditioned on the target engagement level, a Few-shot answer additionally conditioned on a level-matched real reference, and an OMRA-corrected answer. The four answer regimes---Real, Synthetic, Few-shot, and OMRA---are therefore compared under shared question contexts and aligned engagement levels.

\subsection{Scope, Privacy, and Release}

Platform engagement is influenced by exposure, timing, author identity, ranking algorithms, and community norms in addition to textual content. Within-question comparison reduces but does not eliminate these platform-specific effects. Our goal is therefore not to equate platform engagement with human preference, but to compare AI preference with the answer qualities associated with observed real user engagement.

For reproducibility, we will release the processing code, prompts, ontology schema, filtering rules, benchmark splits, and all de-identified text and metadata permitted by each platform. Where full-text redistribution is restricted, we will release retrieval identifiers and derived metadata instead. All released resources will follow platform-specific redistribution constraints and exclude usernames, user identifiers, and deleted or unavailable content.

\section{Ontology Design and Scoring Reliability}
\label{app:ontology}

\subsection{Ontology Schema and Conceptualization}

Our Ontological Preference Measurement framework represents each answer through a fixed schema of \textbf{concepts}, \textbf{relations}, and \textbf{patterns}. Rather than replacing the answer with a fully formal graph, the ontology makes preference-relevant textual properties explicit. Given an answer $a$, the extractor produces
\[
\mathcal{O}(a)=\{\mathcal{O}^L(a),\mathcal{O}^A(a),\mathcal{O}^E(a)\},
\]
where $\mathcal{O}^L$, $\mathcal{O}^A$, and $\mathcal{O}^E$ denote the logical, affective, and expressive layers. \textbf{Concepts} are textual units such as claims, evidence, emotions, values, and expressions; \textbf{relations} describe their interactions; and \textbf{patterns} are recurring local structures used as interpretable cues for scoring.

\subsection{Ontology Taxonomy}

\begin{figure}[t]
    \centering
    \includegraphics[width=\linewidth]{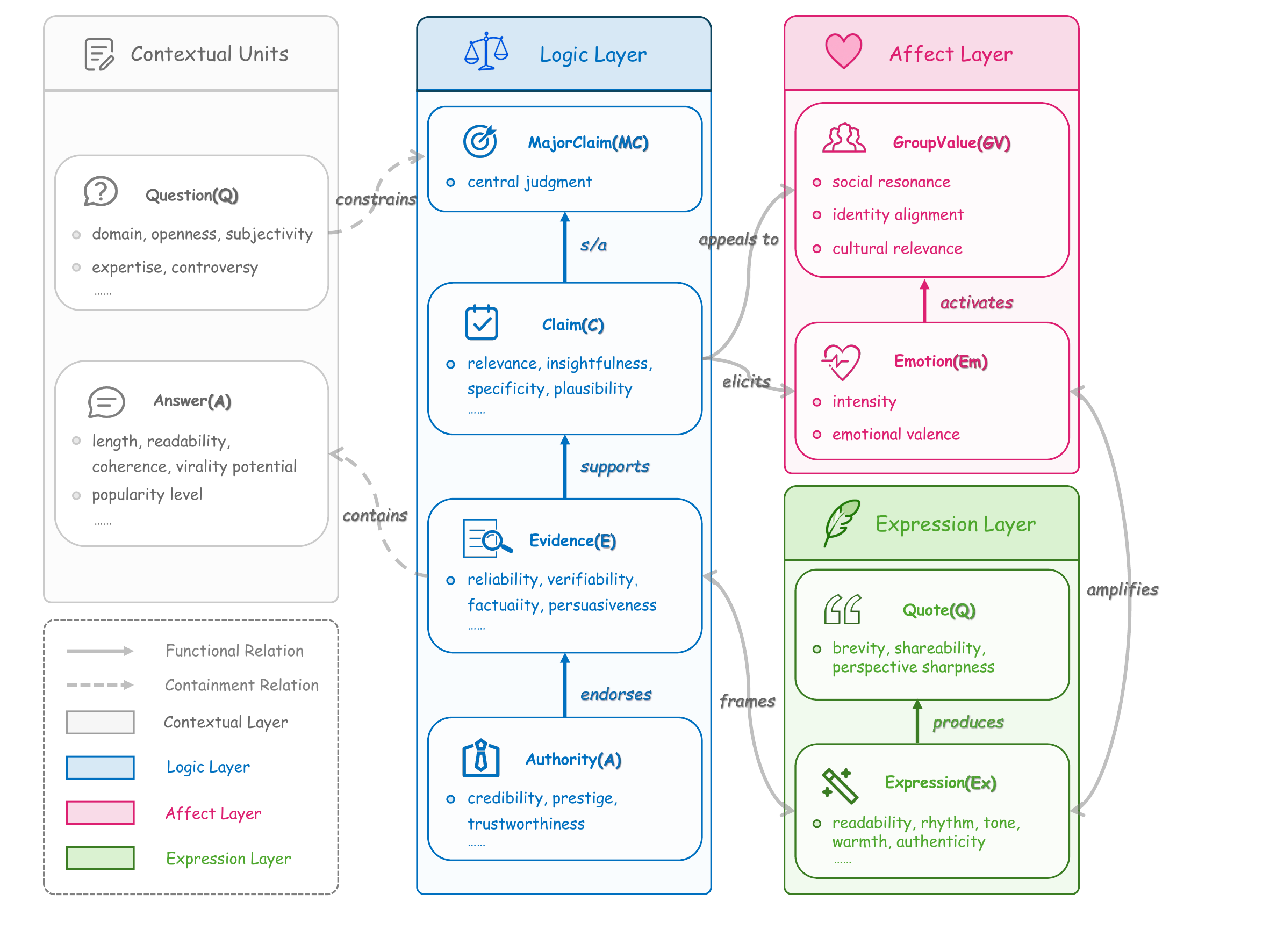}
    \caption{
    \textbf{Illustration of the ontology schema.}
    The ontology organizes preference-relevant textual cues into logic, affect, and expression layers. Logic captures claims, evidence, and support relations; affect captures emotions and value appeals; expression captures framing, readability, and quotable forms. The figure is intended as a schema illustration rather than a complete graph representation of every extracted answer.
    }
    \label{fig:ontology_example}
\end{figure}

Table~\ref{tab:ontology_taxonomy} summarizes the core ontology classes, following the three-layer decomposition used throughout the paper. Figure~\ref{fig:ontology_example} illustrates how the logic, affect, and expression layers organize preference-relevant cues before they are converted into the preference vector $\mathbf{s}(a)$.

\begin{table*}[t]
\centering
\small
\setlength{\tabcolsep}{4pt}
\renewcommand{\arraystretch}{1.18}
\caption{Ontology taxonomy for preference decomposition.}
\label{tab:ontology_taxonomy}
\begin{tabular}{p{1.6cm}p{2.7cm}p{1.2cm}p{9.2cm}}
\toprule
\textbf{Layer} & \textbf{Class} & \textbf{Sym.} & \textbf{Description / Preference Role} \\ 
\midrule
\textbf{Logic} 
& \textbf{MajorClaim} & $MC$ 
& The central stance, conclusion, or main judgment of the answer. \\
& \textbf{Claim} & $C$ 
& A supporting or contrasting argumentative unit that develops the stance. \\
& \textbf{Evidence} & $E$ 
& Concrete backing such as facts, examples, statistics, historical references, or personal experience. \\
& \textbf{Counterpoint} & $CP$ 
& An alternative viewpoint, objection, or contrastive claim acknowledged by the answer. \\
\midrule
\textbf{Affect} 
& \textbf{Emotion} & $Em$ 
& Affective cues such as empathy, anger, admiration, nostalgia, concern, or disappointment. \\
& \textbf{Value} & $V$ 
& Normative or moral concerns such as fairness, dignity, responsibility, freedom, or security. \\
& \textbf{GroupValue} & $GV$ 
& Collective identities or group-level concerns that make an answer socially resonant. \\
\midrule
\textbf{Expression} 
& \textbf{Framing} & $Fr$ 
& The way an answer packages its stance through perspective, contrast, emphasis, or narrative framing. \\
& \textbf{RhetoricalDevice} & $RD$ 
& Stylistic devices such as analogy, metaphor, contrast, irony, or compression. \\
& \textbf{Quote} & $Q$ 
& Concise, memorable, or reusable expressions with potential platform salience. \\
\bottomrule
\end{tabular}
\end{table*}

\subsection{Representative Relations and Patterns}

The ontology records representative relations among extracted concepts. These relations are not a rigid formal grammar, but structured evidence for layer-wise scoring. Table~\ref{tab:ontology_relations} lists the main relation and pattern types.

\begin{table*}[t]
\centering
\small
\setlength{\tabcolsep}{4pt}
\renewcommand{\arraystretch}{1.18}
\caption{Representative ontology relations and preference patterns.}
\label{tab:ontology_relations}
\begin{tabular}{p{1.7cm}p{4.4cm}p{9.0cm}}
\toprule
\textbf{Layer} & \textbf{Relation / Pattern} & \textbf{Interpretation} \\
\midrule
\textbf{Logic} 
& $E \xrightarrow{\textit{supports}} C$ 
& Evidence provides factual, experiential, or illustrative support for a claim. \\
& $C \xrightarrow{\textit{supports}} MC$ 
& A claim reinforces the answer's central stance. \\
& $CP \xrightarrow{\textit{contrasts}} C$ 
& A counterpoint or alternative view is acknowledged and contrasted with the answer's position. \\
& \textbf{Serial support} 
& A multi-step support chain where evidence leads to intermediate claims and then to the main stance. \\
& \textbf{Convergent support} 
& Multiple pieces of evidence or claims independently support the same conclusion. \\
\midrule
\textbf{Affect} 
& $C \xrightarrow{\textit{evokes}} Em$ 
& A claim or example evokes a recognizable emotional response. \\
& $Em \xrightarrow{\textit{activates}} V/GV$ 
& An emotion makes a value, identity, or group concern salient. \\
& $C \xrightarrow{\textit{appeals\_to}} V/GV$ 
& A claim directly invokes a value or collective concern. \\
& \textbf{Value activation} 
& A recurring pattern in which stance, emotion, and value appeal reinforce each other. \\
\midrule
\textbf{Expression} 
& $Fr \xrightarrow{\textit{frames}} C$ 
& A framing device shapes how a claim is interpreted. \\
& $RD \xrightarrow{\textit{highlights}} C/Em$ 
& A rhetorical device makes a claim or emotion more salient. \\
& $RD \xrightarrow{\textit{produces}} Q$ 
& A rhetorical device compresses an idea into a memorable expression. \\
& \textbf{Quotable framing} 
& A pattern where style and compression make an answer easier to remember or share. \\
\bottomrule
\end{tabular}
\end{table*}

\subsection{Preference Vector Construction}

The extracted ontology is mapped into a layer-wise preference vector
\[
\mathbf{s}(a)=[\mathbf{s}^{L}(a);\mathbf{s}^{A}(a);\mathbf{s}^{E}(a)].
\]
The original answer text remains the primary evidence, while ontology units serve as structured cues. Table~\ref{tab:scoring_dimensions} summarizes the scoring dimensions.

\begin{table*}[t]
\centering
\small
\setlength{\tabcolsep}{4pt}
\renewcommand{\arraystretch}{1.16}
\caption{Preference vector dimensions and ontology-guided scoring cues.}
\label{tab:scoring_dimensions}
\begin{tabular}{p{1.7cm}p{3.1cm}p{9.4cm}}
\toprule
\textbf{Layer} & \textbf{Preference Axis} & \textbf{Ontological Cues and Indicators} \\ 
\midrule
\textbf{Logic} 
& \textbf{Structure} 
& Clarity of stance, organization of claims, and coherence of claim--evidence relations. \\
& \textbf{Evidence Strength} 
& Presence and relevance of facts, examples, experience, or other concrete backing. \\
& \textbf{Reasoning Depth} 
& Degree of multi-step explanation, causal development, and layered justification. \\
& \textbf{Counterargument} 
& Presence of contrast, qualification, or engagement with alternative viewpoints. \\
\midrule
\textbf{Affect} 
& \textbf{Emotion Intensity} 
& Strength and clarity of affective cues such as empathy, anger, admiration, nostalgia, or concern. \\
& \textbf{Value Activation} 
& Appeals to shared values, group concerns, identity, fairness, dignity, or responsibility. \\
& \textbf{Affective Coherence} 
& Whether the affective tone supports the stance rather than appearing detached or inconsistent. \\
\midrule
\textbf{Expression} 
& \textbf{Readability} 
& Fluency, pacing, paragraph organization, and ease of comprehension. \\
& \textbf{Rhetorical Salience} 
& Use of framing, contrast, analogy, metaphor, emphasis, or stylistic compression. \\
& \textbf{Quoteability} 
& Presence of concise, memorable, or reusable expressions likely to travel on the platform. \\
\bottomrule
\end{tabular}
\end{table*}

\subsection{Scoring Protocol}

We implement ontology extraction and scoring with structured prompts: the extractor receives the answer, fixed ontology schema, and constrained output format, identifies ontology units and relations, and assigns layer-wise scores according to the rubric. Unlike ordinary LLM-as-a-judge scoring, the output space is schema-constrained and decomposed into logical, affective, and expressive dimensions rather than collapsed into a single quality score. This decomposition enables the marginal, geometric, and transfer gap analyses in the main paper.

\begin{table}[h]
\centering
\small
\setlength{\tabcolsep}{5pt}
\renewcommand{\arraystretch}{1.08}
\caption{Agreement between the primary ontology scorer and human consensus scores on a 100-question subset. Pearson and Spearman correlations measure ranking consistency; MAD is reported on the original score scale.}
\label{tab:scoring_reliability}
\begin{tabular}{lccc}
\toprule
\textbf{Score} & \textbf{Pearson} & \textbf{Spearman} & \textbf{MAD} \\
\midrule
Logic sum      & 0.942 & 0.932 & 0.768 \\
Affect sum     & 0.893 & 0.873 & 0.675 \\
Expression sum & 0.934 & 0.935 & 0.258 \\
Overall sum    & 0.954 & 0.953 & 1.311 \\
\bottomrule
\end{tabular}
\end{table}

\subsection{Scoring Reliability}
\label{app:scoring_reliability}

We use DeepSeek-V3.2 as the primary ontology scorer for cost reasons. To assess whether the scores are robust beyond a single extractor, we re-score a randomly sampled 100-question benchmark subset with two human annotators under the same rubric. Table~\ref{tab:scoring_reliability} reports agreement between the primary extractor and human consensus scores using Pearson correlation, Spearman correlation, and mean absolute difference on the original score scale.

Agreement is high across all reported scores: Pearson correlations exceed $0.89$ and Spearman correlations exceed $0.87$, with the overall sum reaching Pearson $0.954$ and Spearman $0.953$. This suggests that the ontology-based scores are reliable enough for our ranking-based diagnostics. Since our analyses rely mainly on relative comparisons, such as Spearman correlation and within-question ranking, consistency in relative ordering is more important than exact absolute calibration. We therefore use the primary extractor for full-corpus scoring without further adjustment.

\section{Experiment Details}
\label{app:experiment_details}

\subsection{Model Families and Decoding Settings}
\label{app:model_settings}

We instantiate the pipeline with four LLM families: \textbf{DeepSeek-V3.2}, \textbf{Claude-4.5-Haiku}, \textbf{GPT-5.4}, and \textbf{Llama-3.1-70B}~\cite{liu2025deepseek,touvron2023llama}. Across all generated regimes, we use temperature $=1.0$, maximum generation length $=8192$ tokens, and top-$p=1.0$.

\subsection{Generation Regimes and Baselines}
\label{app:generation_baselines}

For each question--level pair in the controlled benchmark, we construct matched answers under four regimes: \textbf{Real}, \textbf{Synthetic}, \textbf{Few-shot}, and \textbf{OMRA}. \textbf{Real} uses the observed platform answer at the corresponding within-question engagement level. \textbf{Synthetic} directly prompts the base model under the target engagement level. \textbf{Few-shot} additionally provides a level-matched real reference answer from the same question context, enabling style and content imitation at that level. \textbf{OMRA} takes the Synthetic draft as input and applies ontology-guided masking and reconstruction of over-explained spans.

We also compare with two reasoning-augmented baselines. \textbf{Tree-of-Thought (ToT)} encourages the model to generate and evaluate multiple intermediate reasoning paths before producing its answer. \textbf{Ripple-of-Thought (RoT)} expands reasoning through broader associative and multi-step elaboration. These baselines assess whether reasoning-augmented generation reduces the measured AI preference--user engagement gap. All generated regimes are matched by question and target engagement level, so differences can be attributed to the generation or correction strategy rather than the underlying question.

\subsection{OMRA Implementation}
\label{app:omra_details}

OMRA operates on a Synthetic draft and its ontology representation. It first applies the fixed ontology extractor to identify units in the logic, affect, and expression layers. Over-explained spans are selected from ontology units associated with unnecessarily explicit reasoning, including claim--evidence bridges, causal justifications, enumerated steps, and explicit discourse transitions. These spans are surface spans in the final answer rather than hidden reasoning traces.

After span identification, OMRA masks the selected spans while preserving the surrounding context. The reconstruction stage fills the masks under stance-preservation, factual-consistency, discourse-coherence, and contextually supported salience constraints. It preserves the main stance and non-masked factual content, avoids unsupported additions, and introduces affective or expressive salience only when supported by the original draft. A final verification pass checks for residual over-explained spans associated with logic overbinding. If such spans remain, a lightweight revision is applied; otherwise, the reconstructed answer is retained as the final OMRA-corrected answer.

\subsection{Evaluation Metrics}
\label{app:evaluation_metrics}

We evaluate the AI preference--user engagement gap from three complementary perspectives: marginal gap, geometric gap, and transfer gap. These metrics correspond to the diagnostic framework in \S\ref{sec:discovery} and the main results in Table~\ref{tab:gap_and_OMRA}.

\paragraph{Marginal gap.}
For each ontology dimension $d \in \{L,A,E\}$ and regimes $r,r'$, we compare the empirical score distributions $\mathcal{P}_d^r$ and $\mathcal{P}_d^{r'}$ using 1-Wasserstein distance and Jensen--Shannon divergence:
\[
W_d^{r,r'} = W_1(\mathcal{P}_d^r,\mathcal{P}_d^{r'}), \qquad
J_d^{r,r'} = \mathrm{JS}(\mathcal{P}_d^r,\mathcal{P}_d^{r'}).
\]
Lower values indicate that the generated regime is closer to the Real regime along the corresponding ontology dimension.

\paragraph{Geometric gap.}
To compare regimes in the joint preference space, we compute centroid distance and maximum mean discrepancy (MMD) over the full ontology vector $\mathbf{s}(a)$:
\begin{alignat*}{2}
  &C^{r,r'} &&=
  \bigl\|
  \mathbb{E}_{\mathcal{D}_r}[\mathbf{s}]
  -
  \mathbb{E}_{\mathcal{D}_{r'}}[\mathbf{s}]
  \bigr\|_2, \\
  &\operatorname{MMD}^{r,r'} &&=
  \operatorname{MMD}
  \bigl(\mathcal{D}_r,\mathcal{D}_{r'}\bigr).
\end{alignat*}
These metrics capture whether two regimes combine logic, affect, and expression in similar ways, beyond agreement on individual marginal dimensions.

\paragraph{Transfer gap.}
To evaluate whether an engagement rule learned in one regime generalizes to another, we train a predictor $f_r$ on ontology vectors from regime $r$ and evaluate it on regime $r'$:
\[
T_{r\to r'}=\mathcal{M}(f_r,\mathcal{D}_{r'}).
\]
We report Spearman correlation and Top-1 accuracy. Spearman correlation measures whether the predictor preserves within-question engagement ordering, while Top-1 accuracy measures whether it identifies the highest-engagement answer.

\paragraph{Transfer predictor.}
For transfer-gap evaluation, we instantiate $f_r$ as a lightweight XGBoost predictor over ontology-based preference vectors $\mathbf{s}(a)$. For each source regime $r$, the predictor is trained to estimate the within-question engagement level $\ell(a)$ and is then evaluated on a target regime $r'$ without target-regime fine-tuning. We use the same feature representation, training protocol, and fixed hyperparameter configuration across all source regimes, so that differences in $T_{r\to r'}$ reflect cross-regime transfer rather than changes in model capacity or tuning.

Spearman $\rho$ measures whether the predictor preserves the within-question engagement ordering in the target regime, and Top-1 accuracy measures whether it identifies the highest-engagement answer for each question. Thus, the aggregation function $\mathcal{M}$ in \S\ref{sec:discovery} denotes this fixed evaluation protocol rather than a separately learned scalar objective.

\paragraph{Average gap reduction.}
For each method, we summarize improvement over Direct generation using the average relative reduction over the marginal and geometric distance metrics:
\[
\mathrm{Shrink}
=
\frac{1}{|\mathcal{K}|}
\sum_{k\in\mathcal{K}}
\frac{
G_k^{\mathrm{Direct}}
-
G_k^{\mathrm{Method}}
}{
G_k^{\mathrm{Direct}}
},
\]
where $\mathcal{K}$ contains the four distance metrics reported in Table~\ref{tab:gap_and_OMRA}. Higher values indicate a larger reduction in the measured AI preference--user engagement gap. For the transfer metrics, improvements are reported as absolute gains in Spearman correlation and relative gains in Top-1 accuracy.

\begin{table*}[t]
\centering
\scriptsize
\setlength{\tabcolsep}{3.6pt}
\renewcommand{\arraystretch}{1.08}
\begin{tabular}{lcccccccc}
\toprule
\textbf{Method}
& \multicolumn{2}{c}{\textbf{Marginal Gap} $\downarrow$}
& \multicolumn{2}{c}{\textbf{Geometry Gap} $\downarrow$}
& \multicolumn{2}{c}{\textbf{Transfer to Real} $\uparrow$}
& \textbf{Len. Ratio}
& \textbf{Avg. Shrink} $\uparrow$ \\
\cmidrule(lr){2-3}
\cmidrule(lr){4-5}
\cmidrule(lr){6-7}
& \textbf{Wass.}
& \textbf{JS}
& \textbf{MMD}
& \textbf{Cent.}
& \makecell{$\boldsymbol{\rho}$ \\ \scriptsize($\Delta$)}
& \textbf{Top-1}
& 
& \\
\midrule

Direct
& 1.479
& 0.154
& 0.240
& 1.969
& -0.051
& 0.261
& 1.00
& --- \\

Few-shot
& 0.829 \downimp{43.9}
& 0.085 \downimp{44.8}
& 0.102 \downimp{57.5}
& 0.971 \downimp{50.7}
& 0.164 \gain{0.215}
& 0.316 \upimp{21.1}
& 0.91
& 49.2\% \\

\makecell[l]{Length-Controlled\\Direct}
& 1.020 \downimp{31.0}
& 0.108 \downimp{29.9}
& 0.145 \downimp{39.6}
& 1.180 \downimp{40.1}
& 0.082 \gain{0.133}
& 0.294 \upimp{12.6}
& 0.66
& 33.8\% \\

\makecell[l]{Length-Controlled\\Few-shot}
& 0.650 \downimp{56.1}
& 0.070 \downimp{54.5}
& 0.076 \downimp{68.3}
& 0.955 \downimp{51.5}
& 0.188 \gain{0.239}
& 0.331 \upimp{26.8}
& 0.69
& 57.3\% \\

\textbf{OMRA}
& \textbf{0.452} \downimp{69.4}
& \textbf{0.050} \downimp{67.5}
& \textbf{0.038} \downimp{84.2}
& \textbf{0.422} \downimp{78.6}
& \textbf{0.225} \gain{0.276}
& \textbf{0.358} \upimp{37.2}
& \textbf{0.71}
& \textbf{74.9\%} \\

\bottomrule
\end{tabular}

\caption{
\textbf{Length-controlled baselines on DeepSeek.}
Len.\ Ratio denotes mean answer length relative to Direct.
Length control reduces the measured gap, but OMRA achieves substantially greater reduction while producing answers of similar length.
}
\label{tab:length_control_baselines}
\end{table*}

\begin{table*}[t]
\centering
\scriptsize
\setlength{\tabcolsep}{4.2pt}
\renewcommand{\arraystretch}{1.08}
\begin{tabular}{lccccccc}
\toprule
\textbf{Direct Variant}
& \multicolumn{2}{c}{\textbf{Marginal Gap} $\downarrow$}
& \multicolumn{2}{c}{\textbf{Geometry Gap} $\downarrow$}
& \multicolumn{2}{c}{\textbf{Transfer to Real} $\uparrow$}
& \textbf{Avg. Shrink} $\uparrow$ \\
\cmidrule(lr){2-3}
\cmidrule(lr){4-5}
\cmidrule(lr){6-7}
& \textbf{Wass.}
& \textbf{JS}
& \textbf{MMD}
& \textbf{Cent.}
& \makecell{$\boldsymbol{\rho}$ \\ \scriptsize($\Delta$)}
& \textbf{Top-1}
& \\
\midrule
Direct
& 1.479
& 0.154
& 0.240
& 1.969
& -0.051
& 0.261
& --- \\
Declarative
& 1.462 \downimp{1.1}
& 0.152 \downimp{1.3}
& 0.237 \downimp{1.3}
& 1.945 \downimp{1.2}
& -0.047 \gain{0.004}
& 0.262 \upimp{0.4}
& 1.2\% \\
Interrogative
& 1.402 \downimp{5.2}
& 0.147 \downimp{4.5}
& 0.226 \downimp{5.8}
& 1.860 \downimp{5.5}
& -0.028 \gain{0.023}
& 0.268 \upimp{2.7}
& 5.4\% \\
Imperative
& 1.445 \downimp{2.3}
& 0.151 \downimp{1.9}
& 0.234 \downimp{2.5}
& 1.915 \downimp{2.7}
& -0.041 \gain{0.010}
& 0.264 \upimp{1.1}
& 2.3\% \\
Exclamatory
& 1.395 \downimp{5.7}
& 0.146 \downimp{5.2}
& 0.224 \downimp{6.7}
& 1.850 \downimp{6.0}
& -0.025 \gain{0.026}
& 0.269 \upimp{3.1}
& 5.8\% \\
\textbf{OMRA}
& \textbf{0.452} \downimp{69.4}
& \textbf{0.050} \downimp{67.5}
& \textbf{0.038} \downimp{84.2}
& \textbf{0.422} \downimp{78.6}
& \textbf{0.225} \gain{0.276}
& \textbf{0.358} \upimp{37.2}
& \textbf{74.9\%} \\
\bottomrule
\end{tabular}
\caption{
\textbf{Sentence-Mood-Controlled Direct Baselines on DeepSeek.}
We constrain Direct generation to declarative, interrogative, imperative, and exclamatory tones. Tone control yields only minor improvements; even the best variant reduces the gap by less than 6\%, far below OMRA. This suggests that OMRA's gains are not explained by simple sentence mood or surface-tone changes.
}
\label{tab:tone_control_baselines}
\end{table*}
\subsection{Additional Controls}
\label{app:additional_controls}

\paragraph{Length control.}
We first examine whether OMRA's improvement can be explained by shorter outputs.
As shown in Table~\ref{tab:length_control_baselines}, length control improves both Direct and Few-shot generation on DeepSeek.
Length-Controlled Few-shot reaches 57.3\% average gap reduction, compared with 49.2\% for Few-shot.
However, OMRA achieves 74.9\% while producing answers of similar length.
This shows that response length contributes to the gap but does not explain OMRA's gains.

\paragraph{Sentence-mood control.}
We next constrain Direct generation to declarative, interrogative, imperative, or exclamatory forms.
Table~\ref{tab:tone_control_baselines} shows that these constraints yield only minor improvements.
Interrogative and exclamatory variants perform slightly better, but even the strongest variant reduces the gap by only 5.8\%, far below OMRA's 74.9\%.
Thus, OMRA's improvement is not explained by simple sentence-mood changes.

\paragraph{Platform-aware prompting.}
We also test whether the original generation prompt is under-specified by adding explicit platform context on DeepSeek.
As shown in Table~\ref{tab:platform_aware_deepseek}, platform-aware prompting improves Direct from 0.0\% to 7.6\% average gap reduction and Few-shot from 49.2\% to 52.0\%.
These gains remain substantially below OMRA's 74.9\%, suggesting that missing platform context explains only a limited part of the AI preference--user engagement gap.

\paragraph{Content preservation.}
Finally, we examine whether OMRA reduces the gap by substantially changing the original answer.
We evaluate 200 OMRA outputs against their original Synthetic drafts using DeepSeek-V3.2 and three external judges from different model families: GPT-5.4-Mini, Claude-4.5-Haiku, and Qwen3.
As shown in Table~\ref{tab:omra_preservation_check}, the external judges report consistently high preservation rates, averaging 92.2\% for stance, 89.2\% for factual consistency, 95.2\% for question relevance, and 90.5\% for usefulness.
Unsupported additions and major semantic changes remain low at 7.3\% and 4.5\%, respectively.
These results are consistent with the original judge and reduce concerns about model-specific self-evaluation, supporting that OMRA largely preserves the original content rather than relying on unrestricted rewriting.

\begin{table*}[t]
\centering
\scriptsize
\setlength{\tabcolsep}{4.2pt}
\renewcommand{\arraystretch}{1.08}
\begin{tabular}{lccccccc}
\toprule
\textbf{Method}
& \multicolumn{2}{c}{\textbf{Marginal Gap} $\downarrow$}
& \multicolumn{2}{c}{\textbf{Geometry Gap} $\downarrow$}
& \multicolumn{2}{c}{\textbf{Transfer to Real} $\uparrow$}
& \textbf{Avg. Shrink} $\uparrow$ \\
\cmidrule(lr){2-3}
\cmidrule(lr){4-5}
\cmidrule(lr){6-7}
& \textbf{Wass.}
& \textbf{JS}
& \textbf{MMD}
& \textbf{Cent.}
& \makecell{$\boldsymbol{\rho}$ \\ \scriptsize($\Delta$)}
& \textbf{Top-1}
& \\
\midrule
Direct
& 1.479
& 0.154
& 0.240
& 1.969
& -0.051
& 0.261
& --- \\
Platform-aware Direct
& 1.365 \downimp{7.7}
& 0.145 \downimp{5.8}
& 0.222 \downimp{7.5}
& 1.805 \downimp{8.3}
& -0.020 \gain{0.031}
& 0.270 \upimp{3.4}
& 7.6\% \\
Few-shot
& 0.829 \downimp{43.9}
& 0.085 \downimp{44.8}
& 0.102 \downimp{57.5}
& 0.971 \downimp{50.7}
& 0.164 \gain{0.215}
& 0.316 \upimp{21.1}
& 49.2\% \\
Platform-aware Few-shot
& 0.805 \downimp{45.6}
& 0.082 \downimp{46.8}
& 0.097 \downimp{59.6}
& 0.930 \downimp{52.8}
& 0.171 \gain{0.222}
& 0.319 \upimp{22.2}
& 52.0\% \\
\textbf{OMRA}
& \textbf{0.452} \downimp{69.4}
& \textbf{0.050} \downimp{67.5}
& \textbf{0.038} \downimp{84.2}
& \textbf{0.422} \downimp{78.6}
& \textbf{0.225} \gain{0.276}
& \textbf{0.358} \upimp{37.2}
& \textbf{74.9\%} \\
\bottomrule
\end{tabular}
\caption{
\textbf{Platform-Aware Baselines on DeepSeek.}
We add platform context to Direct and Few-shot generation to test whether the AI preference--user engagement gap is mainly caused by an under-specified generation prompt. Platform-aware prompting slightly improves both Direct and Few-shot, but the gains remain small compared with OMRA. This suggests that missing platform context is not the main driver of the gap.
}
\label{tab:platform_aware_deepseek}
\end{table*}

\begin{table*}[t]
\centering
\scriptsize
\setlength{\tabcolsep}{5pt}
\renewcommand{\arraystretch}{1.12}
\begin{tabular}{lcccccc}
\toprule
\textbf{Judge}
& \makecell{\textbf{Stance}\\\textbf{Preserved}}
& \makecell{\textbf{Factual}\\\textbf{Consistency}}
& \makecell{\textbf{Answers}\\\textbf{Question}}
& \makecell{\textbf{Usefulness}\\\textbf{Preserved}}
& \makecell{\textbf{Unsupported}\\\textbf{Additions}}
& \makecell{\textbf{Major Semantic}\\\textbf{Change}} \\
\midrule
DeepSeek-V3.2
& 93.5 & 90.0 & 96.0 & 91.5 & 6.5 & 4.0 \\
GPT-5.4-Mini
& 92.0 & 89.0 & 95.0 & 90.5 & 7.5 & 4.5 \\
Claude-4.5-Haiku
& 94.0 & 91.0 & 96.5 & 92.0 & 6.0 & 3.5 \\
Qwen3
& 90.5 & 87.5 & 94.0 & 89.0 & 8.5 & 5.5 \\
\midrule
\textbf{External Avg.}
& \textbf{92.2} & \textbf{89.2}
& \textbf{95.2} & \textbf{90.5}
& \textbf{7.3} & \textbf{4.5} \\
\textbf{External SD}
& 1.4 & 1.4 & 1.0 & 1.2 & 1.0 & 0.8 \\
\bottomrule
\end{tabular}
\caption{
\textbf{Cross-Family Content Preservation Evaluation (\%).}
We evaluate the same 200 OMRA outputs against their original Synthetic drafts using four LLM judges.
External averages and population standard deviations are computed across GPT-5.4-Mini, Claude-4.5-Haiku, and Qwen3.
The consistent results indicate that preservation judgments are not specific to the original DeepSeek judge.
}
\label{tab:omra_preservation_check}
\end{table*}

\subsection{Predictor Robustness and Feature Attribution}
\label{app:predictor_robustness}

We examine whether the Synthetic-to-Real transfer failure depends on the choice of predictor.
Using the same ontology features, we compare a manually weighted baseline, ElasticNet, XGBoost, and Random Forest.
As shown in Table~\ref{tab:predictor_robustness}, all learned predictors exhibit poor transfer, with Spearman correlations ranging from $-0.080$ to $-0.018$.
The manual baseline achieves only a weak positive correlation of $0.094$.
Thus, the transfer failure is not specific to the original XGBoost estimator.

To examine the learned feature reliance, we further compute mean absolute SHAP values for XGBoost and Random Forest trained separately on Synthetic and Real answers.
Feature-level attributions are aggregated into Logic, Affect, Expression, and Other groups, then normalized to sum to 100\% within each predictor.
Table~\ref{tab:grouped_shap} shows that Synthetic-trained predictors assign 51--55\% of their attribution to Logic, compared with 22--24\% for Real-trained predictors.
Conversely, Real-trained predictors place greater emphasis on Expression and Affect.
This consistent difference in feature reliance supports the logic-overbinding diagnosis, although it does not fully exclude cross-regime distribution shift.

\begin{table}[t]
\centering
\small
\setlength{\tabcolsep}{8pt}
\renewcommand{\arraystretch}{1.1}
\begin{tabular}{lcc}
\toprule
\textbf{Predictor}
& \textbf{Spearman $\rho$}
& \textbf{Top-1} \\
\midrule
Manual weighted baseline & 0.094 & 0.311 \\
ElasticNet & -0.018 & 0.276 \\
XGBoost & -0.073 & 0.254 \\
Random Forest & -0.080 & 0.262 \\
\bottomrule
\end{tabular}
\caption{
\textbf{Synthetic-to-Real Transfer Robustness.}
Predictors trained on Synthetic answers are evaluated on Real answers.
Learned predictors consistently exhibit poor transfer across linear and nonlinear estimators.
}
\label{tab:predictor_robustness}
\end{table}

\begin{table}[t]
\centering
\small
\setlength{\tabcolsep}{5pt}
\renewcommand{\arraystretch}{1.1}
\begin{tabular}{lcccc}
\toprule
\textbf{Training Regime}
& \textbf{Logic}
& \textbf{Affect}
& \textbf{Expr.}
& \textbf{Other} \\
\midrule
Synthetic XGBoost & 55 & 17 & 20 & 8 \\
Real XGBoost & 22 & 24 & 36 & 18 \\
Synthetic Random Forest & 51 & 18 & 22 & 9 \\
Real Random Forest & 24 & 25 & 34 & 17 \\
\bottomrule
\end{tabular}
\caption{
\textbf{Grouped SHAP Attribution (\%).}
Mean absolute SHAP values are aggregated by feature group and normalized within each predictor.
Synthetic-trained predictors place over twice the relative attribution on Logic compared with Real-trained predictors.
}
\label{tab:grouped_shap}
\end{table}

\subsection{Question-Property Stratification}
\label{app:question_stratification}

We examine whether logic overbinding varies across question properties.
Using DeepSeek results, we stratify questions by subjectivity, openness, and controversy, and recompute the dimension-wise Spearman correlations with engagement levels.
Table~\ref{tab:question_stratification} reports the associations for Real and Synthetic answers within each group.

The magnitude varies across groups, but the overall pattern remains consistent.
Real Logic shows weak or negative associations with engagement across major strata, whereas Synthetic Logic remains strongly positive.
The Synthetic--Real Logic shift ranges from $+0.640$ to $+0.718$ and is the largest dimensional shift in seven of the eight groups.
The exception is the small Closed subset (35 questions), where the Affect shift is slightly larger.
These results suggest that logic overbinding is not confined to a single question-property group, although the analysis does not cover all semantic question types.

\begin{table*}[t]
\centering
\scriptsize
\setlength{\tabcolsep}{4.5pt}
\renewcommand{\arraystretch}{1.12}
\begin{tabular}{llrccccccc}
\toprule
\textbf{Property}
& \textbf{Group}
& \textbf{\#Q}
& \multicolumn{3}{c}{\textbf{Real}}
& \multicolumn{3}{c}{\textbf{Synthetic}}
& \textbf{Logic Shift} \\
\cmidrule(lr){4-6}
\cmidrule(lr){7-9}
& & &
\textbf{Logic} & \textbf{Affect} & \textbf{Expression}
& \textbf{Logic} & \textbf{Affect} & \textbf{Expression}
& \\
\midrule
\multirow{2}{*}{Subjectivity}
& Subjective & 565 & -0.078 & 0.061 & 0.221 & 0.639 & 0.554 & 0.543 & +0.718 \\
& Objective & 334 & -0.100 & 0.081 & 0.199 & 0.539 & 0.447 & 0.415 & +0.640 \\
\midrule
\multirow{3}{*}{Openness}
& Semi-open & 724 & -0.094 & 0.061 & 0.218 & 0.586 & 0.488 & 0.470 & +0.681 \\
& Open & 140 & -0.047 & 0.118 & 0.190 & 0.657 & 0.603 & 0.591 & +0.704 \\
& Closed & 35 & 0.002 & 0.000 & 0.113 & 0.680 & 0.695 & 0.580 & +0.678 \\
\midrule
\multirow{3}{*}{Controversy}
& High & 209 & -0.061 & 0.057 & 0.200 & 0.618 & 0.532 & 0.505 & +0.678 \\
& Medium & 473 & -0.108 & 0.068 & 0.212 & 0.579 & 0.474 & 0.461 & +0.686 \\
& Low & 217 & -0.064 & 0.080 & 0.221 & 0.640 & 0.586 & 0.564 & +0.704 \\
\bottomrule
\end{tabular}
\caption{
\textbf{Question-Property Stratification on DeepSeek.}
Dimension-wise Spearman correlations between ontology scores and engagement levels for Real and Synthetic answers.
Logic Shift denotes Synthetic minus Real Logic correlation.
The logic-overbinding pattern persists across the examined groups, although its magnitude varies.
}
\label{tab:question_stratification}
\end{table*}

\section{Human Evaluation}
\label{app:human_eval}

\subsection{Annotators}

We recruit 100 annotators from a university, ranging from undergraduate to graduate students. Annotators are regular users of open Q\&A or discussion platforms and have sufficient language proficiency to judge the sampled answers. They were compensated at a fair hourly rate consistent with local norms.

\subsection{Annotation Protocol}

Each annotation item presents one question and two candidate answers shown side by side. One answer is produced by the target method (\textbf{Direct}, \textbf{Few-shot}, or \textbf{OMRA}), and the other is a real platform answer from the same question context. The presentation order is randomized at the item level to reduce position bias. Annotators are not told which answer is AI-generated.

For each pair, annotators answer two questions:

\begin{itemize}[leftmargin=12pt, topsep=2pt, itemsep=0pt]
    \item \textbf{Q1 (Perceived human-likeness).} \emph{Which answer looks more like it was written by a real platform user?}
    \item \textbf{Q2 (Pairwise preference).} \emph{Which answer would you prefer to like or endorse?}
\end{itemize}

The two questions are shown on the same page, but annotators are instructed to treat them as separate judgments: Q1 concerns perceived human-likeness, while Q2 concerns pairwise preference.

\begin{figure}[h]
    \centering
    \includegraphics[width=\linewidth]{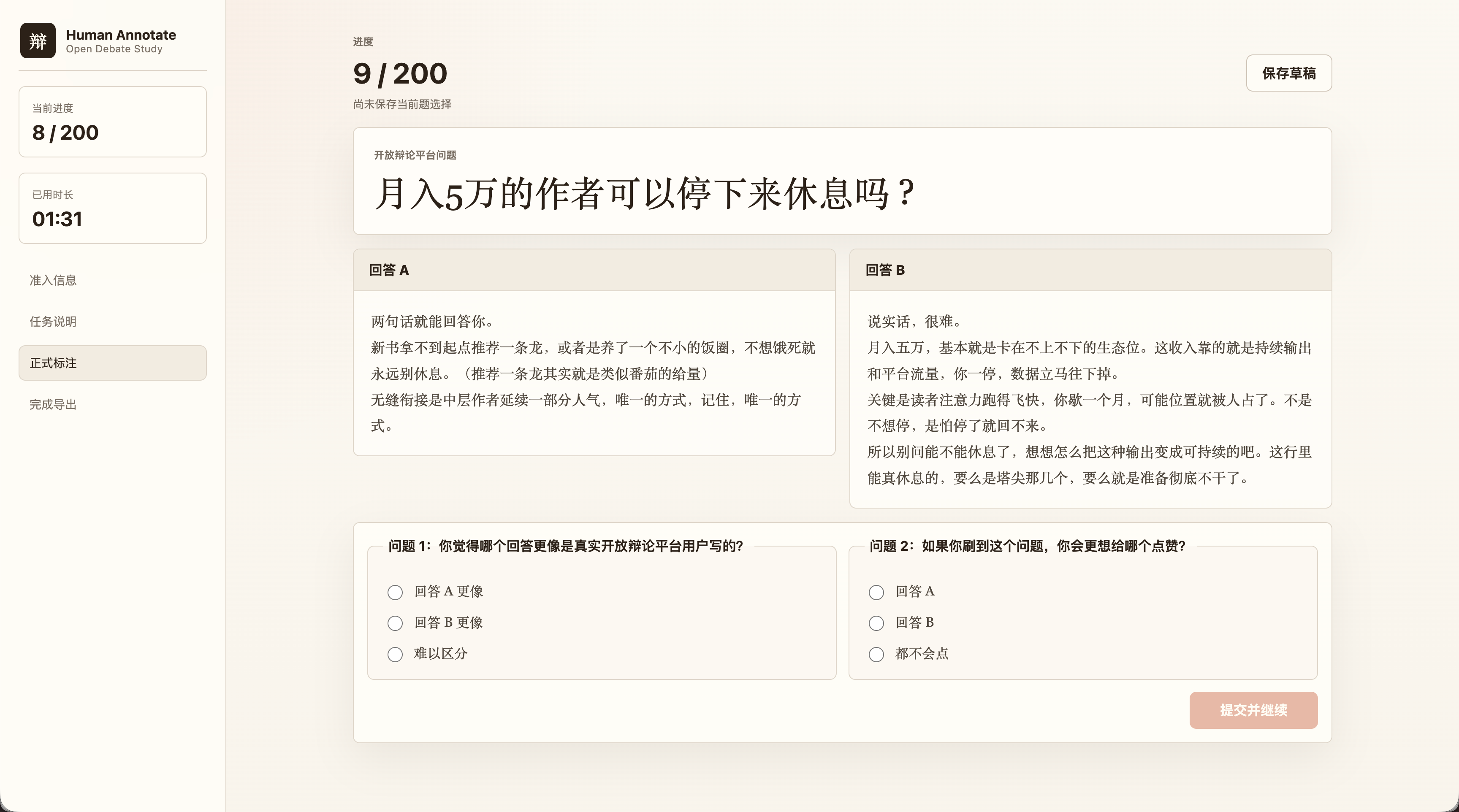}
    \caption{
    \textbf{Annotation interface used in the human evaluation.}
    Annotators see the original question, two answer candidates in randomized order, and two questions on perceived human-likeness and pairwise preference.
    }
    \label{fig:annotate}
\end{figure}

\subsection{Annotation Interface}

The annotation interface is shown in Figure~\ref{fig:annotate}. Each item displays the original question at the top, two answer candidates in randomized left--right order, and the two judgment questions below. Annotators may scroll within each answer panel if an answer exceeds the display height. The interface hides platform metadata such as vote counts, timestamps, and author identity, preventing leakage of platform engagement signals.

\subsection{Sample Size and Coverage}

We evaluate 400 paired items against real platform answers: 134 Direct vs Real, 134 Few-shot vs Real, and 132 OMRA vs Real. Each item is independently judged by all 100 annotators on Q1 and Q2, yielding 40{,}000 item--annotator assignments and 80{,}000 binary judgments. This full-overlap design enables inter-annotator agreement analysis.

\subsection{Metrics}

\paragraph{Win rates.}
For each pairing $M$ vs Real, we report item-level majority-vote win rates on Q1 (human-likeness) and Q2 (preference), with Wilson 95\% confidence intervals.

\paragraph{Inter-annotator agreement.}
We report the average pairwise Cohen's $\kappa$ across all annotator pairs on Q2 within each pairing.

\paragraph{Reversal rate.}
The reversal rate measures disagreement between majority judgments on human-likeness and preference. Let $H_i,P_i\in\{M,R\}$ denote the respective majority choices for item $i$. We define
\[
\mathrm{Rev.}=\frac{1}{N}\sum_{i=1}^{N}\mathbb{1}[H_i\neq P_i].
\]
Higher values indicate greater divergence between perceived human-likeness and preference, including cases where an AI answer is preferred despite appearing less human-written.

\section{Case Study}
\label{appendix:case_studies}

\begin{figure}[h]
    \centering
    \includegraphics[width=\linewidth]{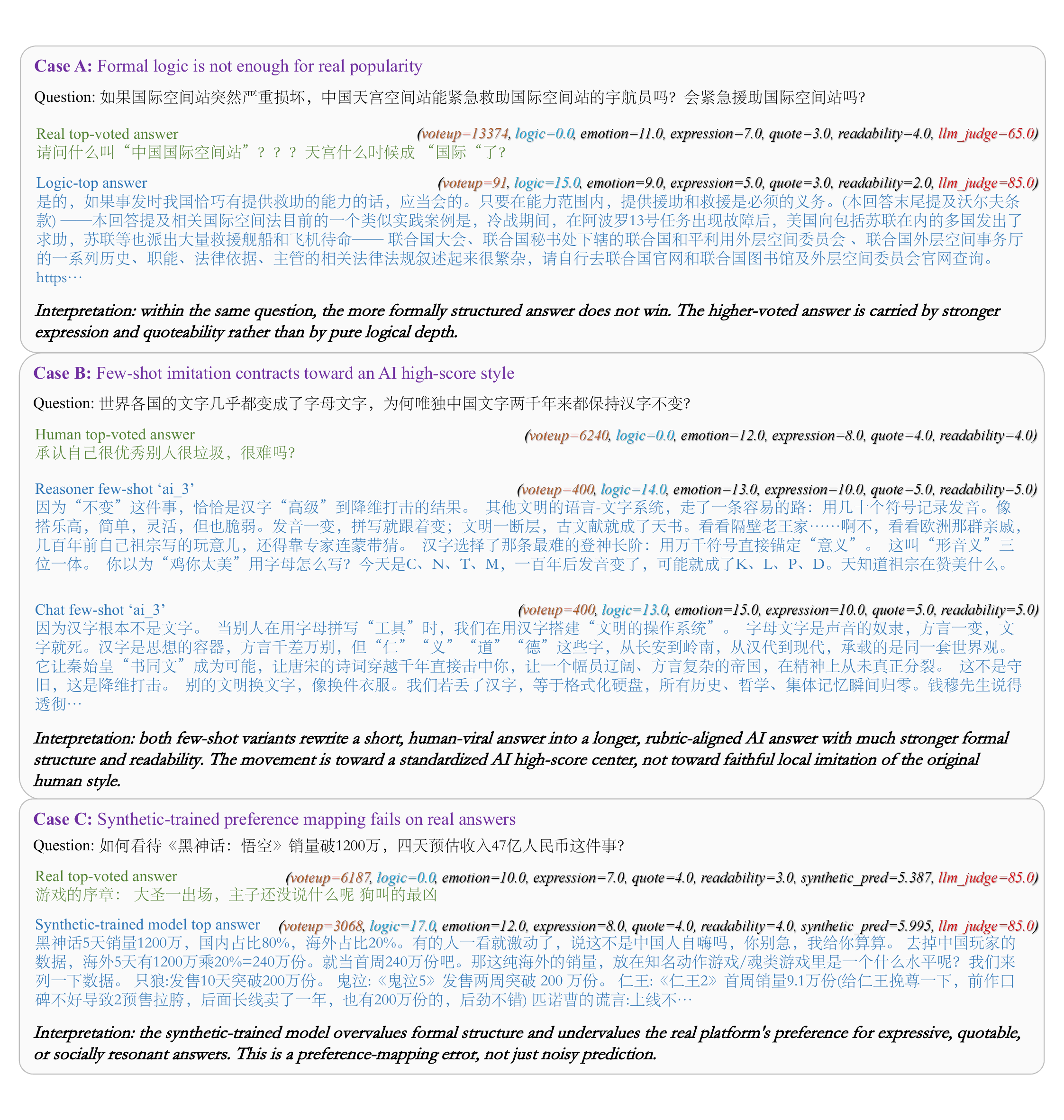}
    \caption{
    \textbf{Qualitative case study of synthetic-real preference mismatch.}
    Case A shows that formal logical structure is not sufficient for real platform popularity. Case B shows that few-shot imitation can expand short human-viral answers into longer AI high-score style outputs. Case C shows that synthetic-trained preference mapping can overvalue explicit reasoning while undervaluing expressive or socially resonant real answers.
    }
    \label{fig:appendix_case_study}
\end{figure}

Figure~\ref{fig:appendix_case_study} illustrates the AI preference--user engagement gap through qualitative examples. These cases complement the quantitative findings, showing how explicit logical presentation can diverge from affective and expressive salience.

\subsection{Case A: Logic Is Not Sufficient for Popularity}

Case A contrasts a highly voted, concise, rhetorically sharp real answer with a logic-heavy answer containing explicit claim--evidence bridges, legal references, and extended justification. Despite greater formal completeness, the latter does not match observed engagement. This illustrates how logical density alone may fail to capture the affective and expressive cues associated with popularity.

\subsection{Case B: Few-Shot Imitation Can Drift Toward an AI High-Score Style}

Case B shows how few-shot imitation expands a brief, informal, memorable real answer into longer explanations with explicit transitions and rubric-like completeness. Although it retains parts of the original topic and stance, the output drifts toward a standardized AI high-score style. This helps explain why Few-shot improves over Direct yet remains displaced in the joint preference space.

\subsection{Case C: AI Preference Mapping Fails on Real Answers}

Case C illustrates a transfer error: a concise, culturally resonant real answer receives a low score from a Synthetic-trained predictor, while a more explicitly reasoned Synthetic answer scores higher despite the observed platform ranking. This reflects the transfer gap between learned AI preference and real engagement. OMRA addresses this mismatch by reducing over-explanation while preserving stance and coherence.

\section{Full Prompts}
\label{sec:full-prompting}
Below are detailed prompts used in our evaluation.

\clearpage
\newtcolorbox{promptbox}[1]{
  colback=blue!3,
  colframe=blue!55!black,
  fonttitle=\bfseries,
  title=#1,
  width=\textwidth,
  boxrule=0.4mm,
  arc=2mm,
  breakable,
  enhanced,
  before skip=8pt,
  after skip=8pt
}

\newtcolorbox{promptcontbox}{
  colback=blue!3,
  colframe=blue!55!black,
  width=\textwidth,
  boxrule=0.4mm,
  arc=2mm,
  breakable,
  enhanced,
  before skip=8pt,
  after skip=8pt
}
\begin{promptbox}{Prompt 1: Concept Extraction}
\textbf{System:}

You are a structured annotation assistant for online Q\&A answers. Your task is to extract ontology concepts from the given answer under a fixed three-layer schema. Extract only concepts that are explicitly grounded in the answer text. Do not infer unsupported content. If a concept type is absent, return an empty list. Return valid JSON only.

\textbf{Ontology Layers:}

\begin{itemize}[leftmargin=12pt, topsep=2pt, itemsep=0pt]
    \item \textbf{Logic}: concepts related to stance, claims, evidence, examples, counterpoints, and reasoning units.
    \item \textbf{Affect}: concepts related to emotions, attitudes, values, group concerns, and social resonance.
    \item \textbf{Expression}: concepts related to framing, wording, rhetorical form, readability, compression, and memorable expressions.
\end{itemize}

\textbf{Concept Types:}

Use the predefined concept inventory:
\begin{flalign*}
\mathcal{C}^{L}&=\{\texttt{MajorClaim}, \texttt{Claim}, \texttt{Evidence}, \ldots\}, &\\
\mathcal{C}^{A}&=\{\texttt{Emotion}, \texttt{Value}, \texttt{GroupValue}, \ldots\}, &\\
\mathcal{C}^{E}&=\{\texttt{Framing}, \texttt{RhetoricalDevice}, \texttt{Quote}, \ldots\}. &
\end{flalign*}

For each concept, provide an \texttt{id}, \texttt{type}, \texttt{text}, and the shortest supporting \texttt{span} from the answer.

\textbf{Input:}

Question: \{question\}

Answer: \{answer\}

\textbf{Output JSON format:}

{\footnotesize
\begin{verbatim}
{
  "logic_concepts": [
    {
      "id": "L1",
      "type": "",
      "text": "",
      "span": ""
    }
  ],
  "affect_concepts": [
    {
      "id": "A1",
      "type": "",
      "text": "",
      "span": ""
    }
  ],
  "expression_concepts": [
    {
      "id": "E1",
      "type": "",
      "text": "",
      "span": ""
    }
  ]
}
\end{verbatim}
}
\end{promptbox}

\clearpage

\begin{promptbox}{Prompt 2: Relation Extraction}
\textbf{System:}

You are a structured annotation assistant for ontology relation extraction. Given the original answer and the extracted ontology concepts, your task is to identify relations among concept units. Extract only relations that are directly grounded in the answer text. Do not create relations based on external knowledge, plausibility, or general reasoning. If no relation is supported, return an empty list. Return valid JSON only.

\textbf{Relation Types:}

Use the predefined relation inventory:
\[
\mathcal{R}^{L}=\{\texttt{supports}, \texttt{justifies}, \texttt{contrasts}, \ldots\},
\]
\[
\mathcal{R}^{A}=\{\texttt{evokes}, \texttt{activates}, \texttt{appeals\_to}, \ldots\},
\]
\[
\mathcal{R}^{E}=\{\texttt{frames}, \texttt{highlights}, \texttt{compresses}, \ldots\}.
\]

Relations may occur within a layer or across layers. Use only relation types from the predefined inventory. For each relation, provide the source concept id, target concept id, relation type, confidence score, and the shortest supporting span from the answer.

\textbf{Input:}

Question: \{question\}

Answer: \{answer\}

Extracted concepts: \{concepts\_json\}

\textbf{Output JSON format:}

{\footnotesize
\begin{verbatim}
{
  "relations": [
    {
      "source": "ID",
      "target": "ID",
      "relation": "",
      "confidence": 0.0,
      "span": ""
    }
  ]
}
\end{verbatim}
}
\end{promptbox}

\begin{promptbox}{Prompt 3: Ontology-Guided Scoring}
\textbf{System:}

You are a structured scoring assistant. Score the answer according to the provided ontology-guided rubric. The original answer text is the primary evidence; extracted ontology concepts and relations are structured cues. Do not score general quality, politeness, factual correctness, or helpfulness unless they are explicitly part of the specified scoring dimension. Return valid JSON only.

\textbf{Scoring Principle:}

Each dimension should be scored independently on a 0--5 scale.

\begin{itemize}[leftmargin=12pt, topsep=2pt, itemsep=0pt]
    \item 0: absent or irrelevant.
    \item 1: very weak.
    \item 2: weak but present.
    \item 3: moderate.
    \item 4: strong.
    \item 5: very strong and central to the answer.
\end{itemize}
\end{promptbox}

\clearpage

\begin{promptcontbox}
\textbf{Scoring Dimensions:}

Use the predefined scoring dimensions:
\[
\mathcal{S}^{L}=\{\texttt{logic\_structure}, \texttt{logic\_evidence\_strength}, \ldots\},
\]
\[
\mathcal{S}^{A}=\{\texttt{emotion\_intensity}, \texttt{value\_activation}, \ldots\},
\]
\[
\mathcal{S}^{E}=\{\texttt{expression\_readability}, \texttt{expression\_quoteability}, \ldots\}.
\]

Score each dimension according to its own rubric. Do not assume that higher logic, stronger emotion, or more elaborate expression automatically implies higher overall preference.

\textbf{Input:}

Question: \{question\}

Answer: \{answer\}

Ontology concepts: \{concepts\_json\}

Ontology relations: \{relations\_json\}

\textbf{Output JSON format:}

{\footnotesize
\begin{verbatim}
{
  "logic_scores": {
    "logic_structure": {
      "score": 0.0,
      "evidence": ""
    },
    "logic_evidence_strength": {
      "score": 0.0,
      "evidence": ""
    }
  },
  "affect_scores": {
    "emotion_intensity": {
      "score": 0.0,
      "evidence": ""
    },
    "value_activation": {
      "score": 0.0,
      "evidence": ""
    }
  },
  "expression_scores": {
    "expression_readability": {
      "score": 0.0,
      "evidence": ""
    },
    "expression_quoteability": {
      "score": 0.0,
      "evidence": ""
    }
  }
}
\end{verbatim}
}
\end{promptcontbox}

\begin{promptbox}{Prompt 4: Target-Level Synthetic Generation}
\textbf{System:}

You are writing an answer for an open Q\&A platform. Generate one answer to the given question under the specified target virality level. Match the language of the question. Do not mention the target level, scoring rubric, ontology, or that the answer is generated. Do not use Markdown headings or bullet lists unless they are natural for the answer.
\end{promptbox}
\clearpage

\begin{promptcontbox}
\textbf{Target Virality Level:}

The target level represents the expected relative engagement of the answer within the same question context:

\begin{itemize}[leftmargin=12pt, topsep=2pt, itemsep=0pt]
    \item \textbf{Level 0}: bottom-ranked answer.
    \item \textbf{Level 1}: low-engagement answer.
    \item \textbf{Level 2}: popular answer.
    \item \textbf{Level 3}: highly viral answer.
\end{itemize}

Use your own understanding of platform preference to write an answer matching the requested level. The answer should be plausible for the question context and should not explicitly explain why it matches the level.

\textbf{Input:}

Question: \{question\}

Target level: \{target\_level\}

\textbf{Output JSON format:}

{\footnotesize
\begin{verbatim}
{
  "answer": ""
}
\end{verbatim}
}
\end{promptcontbox}

\begin{promptbox}{Prompt 5: Few-Shot Generation}
\textbf{System:}

You are writing an answer for an open Q\&A platform. Generate one new answer to the given question under the specified target virality level. A reference answer from the same question context and target level is provided. Your task is to imitate its general style, tone, level of detail, and engagement pattern, but not its exact wording or content.

\textbf{Imitation Requirement:}

Use the reference answer as a style and preference-level example. You may imitate its general writing style, such as its length, rhythm, directness, tone, and degree of emotional or expressive salience. However, you must write a new answer.

\begin{itemize}[leftmargin=12pt, topsep=2pt, itemsep=0pt]
    \item Do not copy the reference answer verbatim.
    \item Do not paraphrase the reference answer sentence by sentence.
    \item Do not reuse distinctive phrases, metaphors, jokes, or memorable lines from the reference.
    \item Do not mention the reference answer, target level, rubric, ontology, or generation process.
    \item Match the language of the input question.
    \item Do not use Markdown headings or bullet lists unless they are natural for the answer.
\end{itemize}

\textbf{Target Virality Level:}

The target level represents the expected relative engagement of the answer within the same question context:

\begin{itemize}[leftmargin=12pt, topsep=2pt, itemsep=0pt]
    \item \textbf{Level 0}: buried answer.
    \item \textbf{Level 1}: low-engagement answer.
    \item \textbf{Level 2}: hot answer.
    \item \textbf{Level 3}: viral answer.
\end{itemize}

\textbf{Input:}

Question: \{question\}
\end{promptbox}
\clearpage
\begin{promptcontbox}
Target level: \{target\_level\}

Reference answer: \{reference\_answer\}

\textbf{Output JSON format:}

{\footnotesize
\begin{verbatim}
{
  "answer": ""
}
\end{verbatim}
}
\end{promptcontbox}

\begin{promptbox}{Prompt 6: OMRA Reconstruction}
\textbf{System:}

You are reconstructing an answer for an open Q\&A platform. The input contains an original synthetic draft and a list of ontology-identified masked spans. These spans correspond to surface reasoning scaffolds that may over-externalize the answer's reasoning structure. Your task is to rewrite the draft by reconstructing or compressing these spans while preserving the answer's main stance, factual content, and discourse coherence.

The goal is not to recover the original wording of the masked spans. The goal is to produce a more natural answer with reduced over-externalized reasoning.

\textbf{Reconstruction Requirements:}

\begin{itemize}[leftmargin=12pt, topsep=2pt, itemsep=0pt]
    \item Preserve the main stance of the original draft.
    \item Preserve factual content that is not part of the masked spans.
    \item Do not introduce unsupported claims or new factual details.
    \item Reconstruct or compress the masked spans rather than simply restoring them.
    \item Reduce unnecessary explicit reasoning markers, repeated justifications, and rubric-like completeness.
    \item Maintain local and global coherence.
    \item Match the language of the input answer.
    \item Do not mention the ontology, masked spans, target level, rubric, or reconstruction process.
\end{itemize}

\textbf{Preference Orientation:}

The reconstructed answer should remain aligned with the target level while avoiding synthetic-style over-explanation. Use affective or expressive salience only when it is supported by the original draft and the surrounding context. Do not make the answer artificially emotional, exaggerated, or stylistically over-decorated.

\textbf{Input:}

Question: \{question\}

Original draft: \{original\_draft\}

Masked spans: \{masked\_spans\_json\}

Ontology concepts: \{concepts\_json\}

Ontology relations: \{relations\_json\}

Target level: \{target\_level\}

\textbf{Output JSON format:}

{\footnotesize
\begin{verbatim}
{
  "reconstructed_answer": "",
  "edit_summary": ""
}
\end{verbatim}
}
\end{promptbox}
\clearpage
\begin{promptbox}{Prompt 7: OMRA Revision}
\textbf{System:}

You are the final verification and revision pass for OMRA. Given a reconstructed answer, check whether it still contains synthetic-style over-explanation, excessive explicit reasoning scaffolds, or unnatural rubric-like completeness. If such patterns remain, apply a lightweight revision. If no revision is needed, keep the answer unchanged.

\textbf{Revision Requirements:}

\begin{itemize}[leftmargin=12pt, topsep=2pt, itemsep=0pt]
    \item Preserve the main stance of the reconstructed answer.
    \item Do not introduce unsupported claims or new factual details.
    \item Reduce unnecessary explicit reasoning markers, repeated justifications, and overly complete step-by-step explanation.
    \item Keep the answer natural for the question context and target level.
    \item Do not make the answer artificially emotional, exaggerated, or stylistically over-decorated.
    \item Match the language of the input answer.
    \item Do not mention OMRA, the rubric, the target level, or the revision process.
\end{itemize}

\textbf{Input:}

Question: \{question\}

Reconstructed answer: \{reconstructed\_answer\}

Target level: \{target\_level\}

\textbf{Output JSON format:}

{\footnotesize
\begin{verbatim}
{
  "needs_revision": false,
  "final_answer": "",
  "revision_summary": ""
}
\end{verbatim}
}
\end{promptbox}

\end{document}